%% file: main.tex
\documentclass[10pt]{article}
\usepackage[preprint]{tmlr}

\usepackage[T1]{fontenc}
\usepackage{amsmath,amssymb}
\usepackage{graphicx}
\graphicspath{{./}}
\usepackage{booktabs}
\usepackage{url}
\usepackage[most]{tcolorbox}
\usepackage{wrapfig}

\usepackage{xcolor}

\usepackage{capt-of}
\usepackage{float}
\makeatletter\let\kernel@addtocurcol\@addtocurcol\makeatother
\usepackage{flafter}
\makeatletter\newcommand{\floatsabovesource}{\let\@addtocurcol\kernel@addtocurcol}\makeatother
\usepackage{microtype}
\usepackage{amsthm}
\usepackage{hyperref}
\hypersetup{hidelinks}
\usepackage[capitalize,noabbrev]{cleveref}

\newtcolorbox[auto counter]{takeaway}{
  enhanced, sharp corners, colback=black!6, colframe=black!6, boxrule=0pt,
  fontupper=\small, before upper={\textbf{Takeaway \thetcbcounter.}\ },
  left=6pt, right=6pt, top=4pt, bottom=4pt, before skip=8pt, after skip=8pt
}

\input{notation}

\title{A Spectral Perspective on Catastrophic Forgetting}

\author{Mahdi Kallel$^{1,*}$ \enspace Nour Jamoussi$^{2,*}$ \enspace Johannes T\"olle$^{3}$ \enspace Marios Kountouris$^{4,\dagger}$ \enspace Carlo D'Eramo$^{1,\dagger}$\\[2pt]
{\small\normalfont\itshape $^{1}$University of W\"urzburg, Germany \quad $^{2}$EURECOM, France \quad $^{3}$University of Luxembourg, Luxembourg\\
$^{4}$DaSCI, University of Granada, Spain}\\[2pt]
{\small\normalfont\ttfamily \{mahdi.kallel, carlo.deramo\}@uni-wuerzburg.de, nour.jamoussi@eurecom.fr, johannes.toelle@uni.lu, mariosk@ugr.es}}

\begin{document}

\maketitle
{\renewcommand{\thefootnote}{\fnsymbol{footnote}}\footnotetext[1]{Corresponding authors.}\footnotetext[2]{Equal supervision.}}

\input{sections/00_abstract}

\input{sections/01_introduction}

\input{sections/02_setup}
\input{sections/03_memory}

\begingroup
\setlength{\parskip}{4pt}
\setlength{\textfloatsep}{9pt plus 2pt minus 1pt}
\input{sections/04_learning}
\input{sections/05_related}

\input{sections/06_conclusion}

\par\endgroup

\makeatletter
\if@accepted
\subsubsection*{Acknowledgments}
Marios Kountouris was partially supported by the European Research Council (ERC) under the European Union's Horizon 2020 research and innovation programme (Grant Agreement No.~101003431) and by the SNS JU under Horizon Europe (Grant No.~101192080, 6G-LEADER).
Nour Jamoussi was supported by the IMT ``Futur, Ruptures \& Impacts'' programme.
Johannes T\"olle acknowledges support from the PCS-GRAPHS project.
\fi
\makeatother

\bibliography{references}
\bibliographystyle{tmlr}

\clearpage
\appendix
\setcounter{topnumber}{2}
\setcounter{bottomnumber}{1}
\setcounter{totalnumber}{3}
\crefalias{section}{appendix}
\crefalias{subsection}{appendix}
\input{appendix/proof}
\clearpage
\input{appendix/b_setups}
\input{appendix/a_spectral_edge}

\input{appendix/c_controls}
\clearpage

\end{document}

%% file: notation.tex
\newcommand{\forget}[2]{F_{#1#2}}

\newcommand{\Pout}{P_U}
\newcommand{\Pin}{P_V}

\newcommand{\dW}{\Delta W}

\newcommand{\persistence}{\rho}

\newcommand{\overlap}{M}

%% file: sections/00_abstract.tex
\begin{abstract}
Neural networks trained on a sequence of tasks can learn each new one well, yet their performance often drops on previously learned tasks. This phenomenon, known as catastrophic forgetting, has been studied extensively, but it remains unclear where earlier learning is stored in a network's weight matrices and why later training disrupts it.
To locate this interference in the weights, we use the Marchenko–Pastur law from random matrix theory to establish a reference range for the singular values of each weight matrix. We call the directions whose singular values lie outside this range the spectral edge. Across control, vision, and language tasks, undoing weight changes that involve the edge raises loss on the newly learned task and partly restores performance on earlier ones, more so than undoing rank-matched bulk-involving changes.
Moreover, we show that many of the edge modes formed during earlier tasks remain in the edge across subsequent task endpoints. Although the majority of singular directions remain in the bulk, a new task's inputs and errors project strongly onto the inherited upper edge, even before training on that task begins.
We find that subsequent weight changes are disproportionately concentrated in the inherited upper edge. Protecting the spectral edge reduces forgetting more than protecting an equal number of bulk directions, while still allowing new tasks to be learned in our continual learning tests.
Memory and new learning thus draw disproportionately on the same spectral edge.
\end{abstract}

%% file: sections/01_introduction.tex
\section{Introduction}
\label{sec:introduction}

Training a neural network on a new task often degrades its performance on previously learned tasks.
This deterioration is known as catastrophic forgetting \citep{mccloskey1989catastrophic,french1999catastrophic} and reflects the stability side of the stability and plasticity dilemma \citep{carpenter1987art,mermillod2013stability}.
Prior work has connected forgetting to interference between tasks, measured as alignment between their gradients \citep{riemer2019learning} or overlap between their neural tangent kernels \citep{doan2021theoretical}, and to the curvature of an earlier task's loss along the weight change made by later training \citep{mirzadeh2020understanding}.
Other studies locate forgetting mainly in deeper layers and their representations \citep{ramasesh2021anatomy}.

Methods for reducing forgetting estimate the importance of individual parameters from gradients and curvature on earlier tasks \citep{kirkpatrick2017overcoming,zenke2017synaptic}, or store their activation subspaces \citep{saha2021gradient}, in order to protect them during later training.
These methods rely on gradients, curvature or activations computed from past data.
Training, however, shapes the weights of the neural network.
Looking into the weights themselves, we identify which directions within individual layers support earlier learning and study their interaction with the gradients of upcoming tasks.

\begin{figure}[t]
\centering
\includegraphics[width=0.95\linewidth]{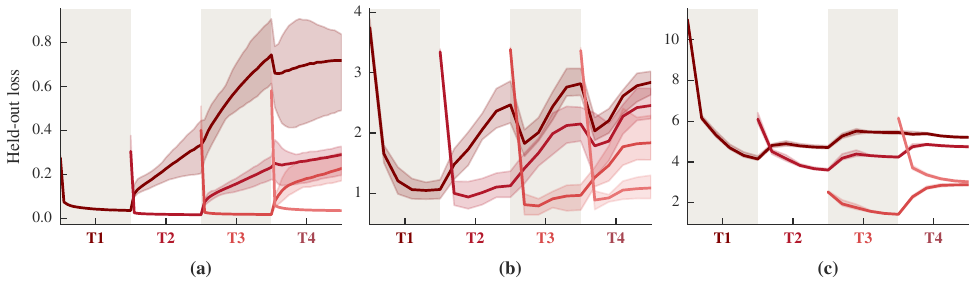}
\caption{\textbf{The loss on earlier tasks rises as later tasks are learned.}
Panels show \textbf{(a)}~the MLP on four MimicGen tasks (T1: square, T2: threading, T3: coffee, T4: three-piece assembly), \textbf{(b)}~ResNet-18 on four disjoint 25-class splits of CIFAR-100 (T1--T4, respectively) and \textbf{(c)}~GPT-2 on four text domains (T1: English, T2: mathematics, T3: code, T4: German). The same task definitions and ordering are used throughout the paper. Each curve shows the validation loss on one task from the start of its own training onward, as the network learns sequentially T1--T4 (darker to lighter red). Lines show the mean over five seeds and bands their range.}
\label{fig:forgetting}
\end{figure}

Singular value decomposition (SVD) breaks down a weight matrix into input and output singular vector pairs whose strength is set by the corresponding singular value.
While optimizing for its objective, training a neural network modifies these strengths and rotates the directions within individual layers.
Spectral methods for updating pretrained models build on this decomposition, treating the directions associated with the largest singular values as the primary carriers of previously acquired knowledge.
Fine-tuning disturbs these directions, as low-rank adaptation adds new dominant directions \citep{shuttleworth2025lora} and supervised fine-tuning rotates the leading ones \citep{zhu2026path}. Moreover, keeping updates out of these dominant directions while leaving the remaining directions free for new learning limits forgetting \citep{xiong2026oplora,nayak2026sculpting,zhang2026spectral}.
Yet these methods fix the protected directions by a chosen rank or energy threshold, and do not explain why later training reaches the directions that hold earlier learning.

We leverage Random Matrix Theory (RMT) to set this threshold in a principled manner, taking the spectrum of a random matrix as a reference for which directions carry learned structure~\citep{wigner1955characteristic,mehta2004random}.
At random initialization, the singular values of a large weight matrix typically lie within a bounded band whose boundaries are predicted by the Marchenko--Pastur (MP) law, commonly called the bulk \citep{marvcenko1967distribution,martin2021implicit,thamm2022random}.
As the network learns, some singular values leave this band, forming what we refer to as the \emph{spectral edge} and revealing structure associated with learning  \citep{martin2021implicit,thamm2022random}.
For example, in pretrained transformers,
this structure forms outside both ends of the bulk, and removing the corresponding directions damages performance more than removing bulk equivalents \citep{staats2025small}.
These studies examine networks trained on a single task or data distribution, leaving open how the weight spectrum, specifically its edge, evolves as a network learns several tasks sequentially and how future learning interacts with the structure left by earlier tasks.

In this work, we identify the spectral edge by fitting the MP bulk to each weight matrix and investigate its evolution throughout sequential learning.
We trace sequential training from scratch and without replay in three settings, where an MLP imitates a sequence of robot manipulation skills, ResNet-18 learns disjoint classes of CIFAR-100, and GPT-2 is trained on a sequence of text domains (\cref{fig:forgetting}).
A pretrained GPT-2 fine-tuned on new text classification tasks extends the study to pretrained networks.

First, the network learns each task disproportionately through the spectral edge (\cref{sec:memory}).
Removing the edge-involving part of a task's weight change  damages that task far more than removing a  rank-matched bulk-involving part. These changes also interfere with earlier learning, since undoing them partly restores earlier tasks. To understand what survives this interference, we follow the edge across tasks. The upper edge keeps most of its subspace from one task to the next, and many of its modes from earlier tasks remain as new ones enter.

Second, later learning lands in the same edge (\cref{sec:learning}).
To understand why new tasks alter these inherited directions, we examine their inputs and errors before training begins. Both already project strongly onto the edge, consistent with the concentration of subsequent weight updates along these directions.
This raises the question whether preventing changes through these directions protects earlier tasks. For this matter, we accumulate edge directions across tasks and project subsequent updates away from them. This protection reduces forgetting more than protecting as many bulk directions, while new tasks still reach losses close to those of ordinary training (\cref{sec:freeze}).

%% file: sections/02_setup.tex
\section{Setup and notation}
\label{sec:setup}

\subsection{Sequential training and forgetting}
\label{sec:forgetting}

Starting from initial weights $\mathbf{W}_0$, random or pretrained, we train a network on a sequence of tasks, without access to the data of earlier tasks.
Let $\mathbf{W}_i$ denote the weights of the whole network at the end of stage $i$ and $L_j$ the validation loss on task $j$.
We define the learning gain on task $j$ as the decrease in its loss during its own training stage, $L_j(\mathbf{W}_{j-1}) - L_j(\mathbf{W}_j)$, which provides a task-specific scale for comparing loss changes across tasks.
For $i>j$, we define the forgetting of task $j$ after stage $i$ as the rise of its loss since it was learned, $\forget{i}{j} = L_j(\mathbf{W}_i) - L_j(\mathbf{W}_j)$.

We study three settings of four tasks T1--T4 each (\cref{app:setups}).
An MLP imitates robot-arm manipulation skills from MimicGen demonstrations \citep{mandlekar2023mimicgen}, and ResNet-18 \citep{he2016deep} learns disjoint class splits of CIFAR-100 \citep{krizhevsky2009learning}, both task-incremental, with one output head per task.
GPT-2 \citep{radford2019language} learns English, mathematical, code and German text from scratch with one next-token predictor throughout \citep{reid2022m2d2,paster2024openwebmath,codeparrot2021clean,penedo2025fineweb2}.
The MLP and ResNet-18 train with SGD and GPT-2 with AdamW, each over five seeds.
A fourth experiment fine-tunes a pretrained GPT-2 on four text classification tasks \citep{wang2023olora} and investigates the retention of knowledge acquired during pretraining (\cref{sec:freeze}).

\subsection{Singular modes and the spectral edge}
\label{sec:edge}

\begin{minipage}[c]{0.54\linewidth}
We analyze the weight matrices of the hidden layers, reading convolutions as matrices acting on input patches (\cref{app:matrix-views}), and leave out the input layer, the task heads and GPT-2's tied embedding and output matrix. To mitigate forgetting arising from changes outside the hidden weight matrices, we hold the input layer \citep{zheng2025spurious}, the normalization scales and all biases outside the task heads \citep{mallya2018packnet} fixed from the second task in runs trained from scratch and throughout pretrained GPT-2 fine-tuning.
Previous work suggests that freezing the input layer can have little effect on learning subsequent tasks \citep{ramasesh2021anatomy}.
For each layer, we write $W_i$ for its weight matrix in $\mathbf{W}_i$, and $W$ when the stage is irrelevant.
Each matrix $W\in\mathbb{R}^{m\times n}$ maps an input $h\in\mathbb{R}^n$ to $Wh$, and has the compact singular value decomposition $W=U\Sigma V^\top=\sum_{k=1}^{p}\sigma_k u_k v_k^\top$ with $p=\min(m,n)$.
The right singular vector $v_k$ selects an input direction, the left singular vector $u_k$ gives the corresponding output direction, and the singular value $\sigma_k$ sets the strength of the map between them.
We call each triple $(\sigma_k,u_k,v_k)$ a mode.
When $m\neq n$, the larger-dimensional space also contains additional null directions, orthogonal to the compact left singular basis for a tall matrix and to the compact right singular basis for a wide matrix.
\end{minipage}
\hfill
\begin{minipage}[c]{0.44\linewidth}
\centering
\includegraphics[width=\linewidth]{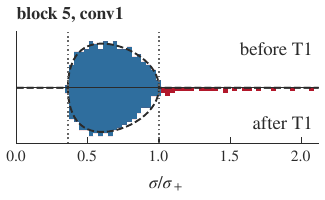}
\setlength{\abovecaptionskip}{4pt}
\captionof{figure}{\textbf{Training forms the spectral edge (ResNet-18).}
Singular values of the first convolution of residual block 5 before (top) and after (bottom) the first task, with the fitted MP density (dashed).
 Most singular values reside in the bulk, close to their distribution at random initialization, while some break out of either end and form the edge. Final spectra of the inspected layers of each network are in \cref{app:spectra-results}.}
\label{fig:spectrum}
\end{minipage}

The Marchenko--Pastur (MP) law predicts the range of singular values of a large random matrix with independent, centered entries \citep{marvcenko1967distribution}.
Neural weight matrices at random initialization are often well described by this model \citep{martin2021implicit,thamm2022random}. A low-rank signal added to a random matrix produces singular values outside the bulk once its strength crosses a threshold \citep{benaych2012singular}. In a solvable model of gradient-flow training, learning appears exactly as such an outlier separating from the bulk \citep{coeurdoux2026random}, and outliers of this kind appear in the weight spectra of trained networks across many architectures \citep{martin2021implicit}.

Following \citet{staats2025small}, we fit the MP bulk separately for each weight matrix at each checkpoint using the median-based scale estimator of \citet{gavish2014optimal} (\cref{app:mp-fit}).
The fitted bulk contains nearly every mode at initialization.
We treat singular directions beyond the fitted bulk at either end as candidate learned structure.
Modes with singular values above the upper cutoff $\sigma_+$ form the upper edge, and those below the lower cutoff $\sigma_-$ form the lower edge. For square matrices, the lower cutoff is zero, so only the upper edge can occur.
Together, these modes form what we call the spectral edge, and the remaining modes form the MP bulk (\cref{fig:spectrum}).

\subsection{Displacements and gradients in the singular basis}
\label{sec:gradients}

We call the net weight change of stage $i$, $\dW_i=W_i-W_{i-1}$, its displacement. Its energy is the sum of the squared weight changes, $\|\dW_i\|_F^2$, where $\|\cdot\|_F$ denotes the Frobenius norm. To relate the displacement to the modes of the layer, we express it in the fixed singular basis of a reference checkpoint. In this basis, each entry $u_k^\top\dW_i v_l$ measures the change in how strongly the layer maps input direction $v_l$ to output direction $u_k$.

To isolate the part of a displacement $\dW$ involving a chosen set of modes, let $P_U$ and $P_V$ be the orthogonal projectors onto the spans of their left and right singular vectors.
The product $(I-P_U)\dW(I-P_V)$ retains only components whose input and output directions both lie outside the selected subspaces.
The part involving the chosen modes is therefore
\begin{equation}
  \dW-(I-P_U)\dW(I-P_V).
  \label{eq:split_edge}
\end{equation}
Choosing the edge modes in \cref{eq:split_edge} gives the \textit{edge-involving} part of $\dW$, and choosing a set of bulk modes gives the \textit{bulk-involving} part. Both include couplings between edge modes and the selected bulk modes, while changes between edge directions belong only to the edge-involving part.

The gradient connects weight changes to the layer's inputs and errors.
For a layer with pre-activation $Wh$, let $\delta$ denote the gradient of the loss with respect to $Wh$, obtained by backpropagation.
By the chain rule, each application of the layer contributes the outer product $\delta h^\top$ to the weight gradient \citep{martens2015optimizing}. The coefficient of this contribution from input direction $v_l$ to output direction $u_k$ is
\begin{equation}
  u_k^\top (\delta h^\top)\, v_l
  = (u_k^\top \delta)(v_l^\top h).
  \label{eq:gradient-coefficients}
\end{equation}
The magnitude of the coefficient depends on the strength of  the input projection onto $v_l$ and the error projection onto $u_k$.
The weight gradient sums these contributions over examples and the positions where the layer is applied, and the optimizer, SGD \citep{robbins1951stochastic} or AdamW \citep{loshchilov2019decoupled} in our settings, converts these gradients into updates whose sum over the stage is the displacement $\dW_i$.

%% file: sections/03_memory.tex
\section{Tasks are learned disproportionately through the edge}
\label{sec:memory}
We examine where inside the weights a task is learned, and what of it persists during training on subsequent tasks.
As the spectral edge emerges from the bulk during training, it provides a natural starting point for this investigation. We first undo the weight changes that involve the edge and measure how the loss on each task is impacted, and then follow the edge itself from one task to the next.

\subsection{Edge changes matter most for learning and forgetting}
\label{sec:fossil}

\begin{figure}[t]
\centering
\includegraphics[width=\linewidth]{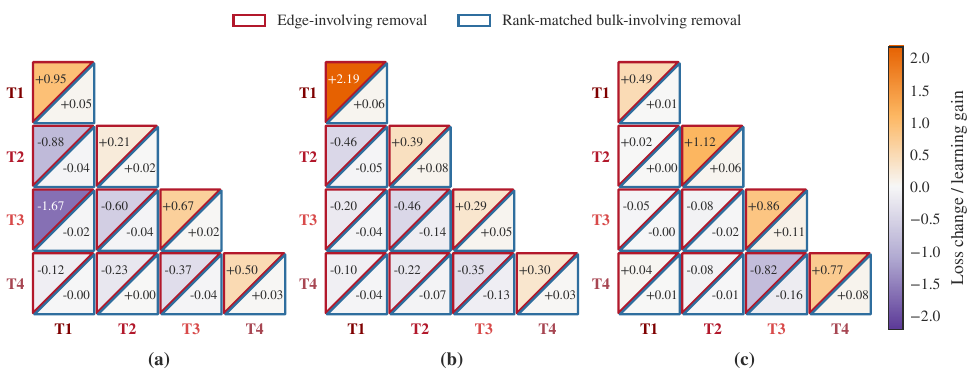}
\caption{\textbf{Removing the edge-involving part of a stage's displacement damages the newly learned task and partly restores earlier tasks.} Panels show \textbf{(a)}~the MLP, \textbf{(b)}~ResNet-18 and \textbf{(c)}~GPT-2.
Rows are training stages and columns evaluated tasks, so the diagonal holds the task learned in that stage.
In each cell, the upper-left triangle removes the edge-involving part and the lower-right triangle the bulk-involving part of as many random bulk modes (\cref{app:removal}).
Values are the change in validation loss in units of the task's learning gain, averaged over five seeds, positive for damage and negative for recovery.
}
\label{fig:fossil}
\end{figure}

\begin{figure}[t]
\centering
\includegraphics[width=\linewidth]{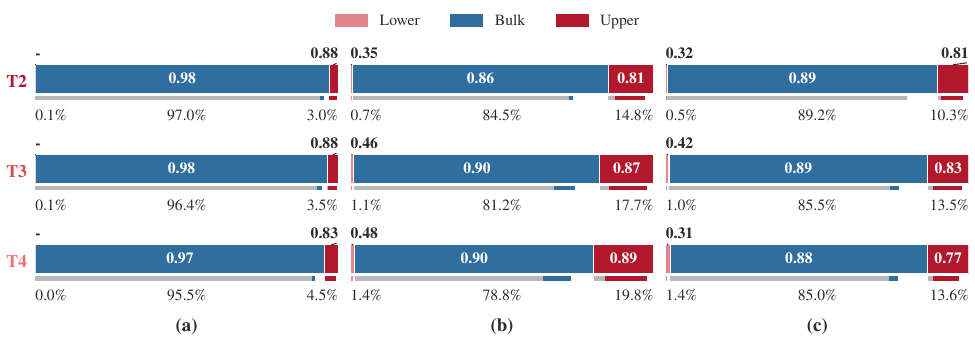}
\caption{\textbf{The upper edge persists across successive tasks.} Panels show \textbf{(a)}~the MLP, \textbf{(b)}~ResNet-18 and \textbf{(c)}~GPT-2, with rows for the stages that learn T2, T3 and T4. Each bar splits the modes after the stage into lower edge, MP bulk and upper edge, with widths and percentages giving each region's share. The number in each region is its persistence across the stage, and a dash marks a region that is empty before or after the stage. The thin bar under each region runs to that persistence on the region's own scale, grey up to its chance level. Values average five seeds. All hidden layers pooled, and the inspected layers of each network, are in \cref{app:persistence-results}.}
\label{fig:stratigraphy}
\end{figure}

If changes involving the edge support what a stage learns, undoing them should impair the task learned in that stage and if a later stage's edge-involving changes drive forgetting, undoing them should partly restore performance on earlier tasks. To test both predictions, we express the displacement $\dW_i$ of stage $i$ in the singular basis of $W_i$, the weights at the end of that stage, whose edge contains the modes the stage has built.
We then remove the edge-involving part of $\dW_i$ from each hidden weight matrix using $P_U^E$ and $P_V^E$, i.e., the projectors associated with the edge modes, in \cref{eq:split_edge} and measure the loss on every task learned so far, keeping the corresponding output head unchanged. As a rank-matched control, we remove the bulk-involving part associated with the same number of randomly chosen bulk modes in each matrix (\cref{app:removal}). This removes changes involving those bulk modes, including their couplings to the edge, while leaving changes between edge directions intact.

Removing the edge-involving part has two effects (\cref{fig:fossil}).
First, it impairs the task learned during that stage more than the rank-matched bulk removal at every stage of every network, by a factor of ten or more on the first task.
Second, removing the edge-involving part of a later task's change lowers the loss on earlier tasks in 16 of 18 comparisons, while the rank-matched bulk removal restores far less.

We rescale the two removals to equal energy to test whether the size of the change explains these effects. At equal energy, edge removal still damages the newly learned task more in 11 of 12 stages and generally leaves earlier tasks with lower loss than bulk removal (\cref{app:energy-results}). Changes that involve the edge thus matter more for what the model learns in each stage, and their removal can substantially alleviate the forgetting of earlier tasks.

\subsection{The edge persists across tasks}
\label{sec:stratigraphy}

A new task's edge changes damage the performance on earlier tasks, but they need not erase the structure those tasks built. To investigate this, we track the edge from one stage to the next.
Each spectral region defines two subspaces, spanned by its left and right singular vectors, respectively. To track a region across a training stage, we compare each of these subspaces before and after the stage.
The alignment between two subspaces can be characterized by their principal angles, i.e., the angles between their corresponding principal vectors, with their squared cosines quantifying their geometric overlap \citep{bjorck1973numerical}.
Let $\mathcal{R}$ and $\mathcal{R}'$ denote the sets of mode indices belonging to a given spectral region before and after a training stage, respectively.
The persistence is defined as the mean squared projection of the original region's singular vectors onto the corresponding subsequent subspaces, averaged over the input and output spaces (\cref{app:continuity}),
\begin{equation}
  \persistence=\frac{1}{2|\mathcal{R}|}\sum_{k\in\mathcal{R}}\sum_{l\in\mathcal{R}'}\Big[(u_k^\top u'_l)^2+(v_k^\top v'_l)^2\Big].
  \label{eq:persistence}
\end{equation}
Equivalently, persistence is the mean squared cosine of the principal angles across both pairs of subspaces, with missing angles contributing zero when the region shrinks (\cref{app:continuity}).
We compute it for the lower edge, the MP bulk and the upper edge separately, whenever the corresponding region is nonempty at both checkpoints.
A persistence of one indicates that both original subspaces are contained in their subsequent counterparts, without requiring individual singular vectors to remain unchanged. Since larger subsequent subspaces capture more of a randomly oriented direction, persistence must be interpreted relative to the dimension of the subsequent region.
As a reference, we use the expected squared projection of a unit vector drawn uniformly from the corresponding compact singular subspace at the subsequent checkpoint, which yields the chance level $|\mathcal{R}'|/\min(m,n)$.

Across the three settings, both edges persist well above their chance levels in the pooled results (\cref{app:persistence-results}).
In the example layers (\cref{fig:stratigraphy}; \cref{app:persistence-results}), the upper edge holds only a small fraction of each matrix's modes, with a chance level of at most 0.20, yet most of it persists from one stage to the next, at 0.77 to 0.89.
The bulk typically persists more in absolute terms, and its persistence lies much closer to the level predicted by its size.
In ResNet-18 and GPT-2, the lower edge is smaller than the upper edge and persists less, at 0.31 to 0.48 in their example layers against chance levels below 0.015. The lower edge thus exhibits lower persistence, even though it remains far above chance.
Previous work shows that small singular components can be important for task performance \citep{hsu2022language}. MiLoRA initializes its low-rank adaptation from these components \citep{wang2025milora}.
Moreover, below the MP bulk, their right singular vectors exhibit substantial overlap with eigenvectors of the input-activation covariance matrix, and removing small singular components after fine-tuning can be more damaging than removing them before fine-tuning \citep{staats2025small}.
These observations motivate the hypothesis that the lower edge carries structure that is useful for the current task but less reusable across tasks, contributing to its lower persistence. The upper-edge subspaces, by contrast, remain largely aligned across successive tasks, even as the loss on earlier tasks increases.

\subsection{The edge accumulates modes across tasks}
\label{sec:esd}
The upper edge can preserve its span even as its constituent singular directions change.
To track individual modes across stages, we match the full sets of singular modes at consecutive checkpoints using a one-to-one Hungarian assignment that maximizes the total overlap between their rank-one maps, $\overlap_{kl}=(u_k^\top u'_l)(v_k^\top v'_l)$ (\cref{app:mode-tracking}).
When both modes in a pair lie in the edge, the later mode inherits the earlier one's task-of-entry label. Modes that enter from the bulk take the current task's label. The matching uses only the

\begin{figure}[h]
\centering
\includegraphics[width=0.95\linewidth]{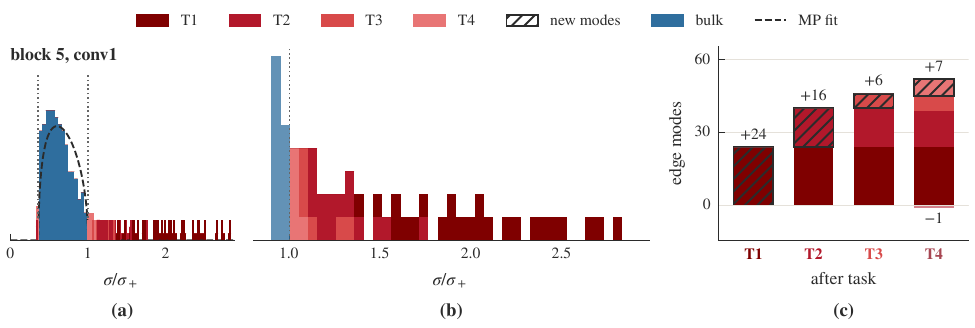}
\caption{\textbf{The edge accumulates modes across tasks, while
earlier modes persist (ResNet-18).} \textbf{(a)}~Spectrum of the first convolution of residual block 5 after the last task, with the fitted MP law and edge modes colored by the task in which they entered the edge. \textbf{(b)}~Zoom into its upper edge, with modes grouped by task of entry. \textbf{(c)}~Edge modes after each task, split into modes that entered in that stage (hatched), modes kept from earlier stages and modes that left for the bulk (below zero). All three panels show the same layer and seed. The same panels for more inspected layers of each network are in \cref{app:spectra-results}.}
 \label{fig:esd}
\end{figure}

\begin{minipage}[c]{0.58\linewidth}
singular vectors, yet each mode and its partner keep nearly the same singular value relative to the upper MP cutoff (\cref{fig:matching-example}; \cref{app:matching}), suggesting that the matching successfully follows a mode across stages.
The assigned pairs let us count modes that enter the edge, stay in it, or leave for the bulk. \cref{fig:esd} illustrates this for ResNet-18 where the colors group the modes by the task in which they entered. More modes typically enter the edge than leave it at each stage.
In the layer of \cref{fig:esd}, all 24 modes that enter the edge with T1 are still in it after T4. Counting every hidden matrix over five seeds, the edge still holds 97\% of its T1 modes after T4 in ResNet-18 and in the MLP, and 81\% in GPT-2, and it grows with every task (\cref{app:spectra-results}).
Later tasks thus start from a larger edge built over several earlier tasks.
\begin{takeaway}
Although the spectral edge contains only a small fraction of each matrix's modes, changes involving it contribute disproportionately to both the learning of new tasks and the forgetting of earlier ones.
Removing the edge-involving part of a stage's displacement damages the task learned in that stage more than the rank-matched bulk removal, and partly restores the tasks learned before it.
Across later stages, most of the upper edge persists and retains many earlier modes as new modes enter the edge, while the lower edge persists less yet still well above chance.
\end{takeaway}
\end{minipage}
\hfill
\stepcounter{footnote}
\begin{minipage}[c]{0.38\linewidth}
\centering
\includegraphics[scale=0.9]{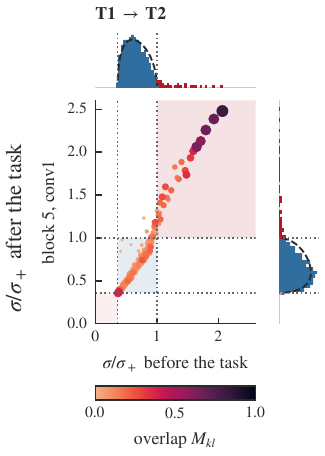}
\setlength{\abovecaptionskip}{4pt}
\captionof{figure}{\textbf{Matched modes keep similar normalized singular values (ResNet-18).} Each mode is placed at its matched partner after T2 and colored by their overlap $\overlap_{kl}$\protect\footnotemark[\value{footnote}], with the spectra before and after the task along the axes (\cref{app:matching}).}
\label{fig:matching-example}
\end{minipage}
\footnotetext{Although positivity is not imposed by the assignment, the matched overlaps are overwhelmingly positive, with only a few negative values close to zero.}

%% file: sections/04_learning.tex
\section{Later learning lands in the same edge}
\label{sec:learning}

Each new task starts from an edge that earlier tasks built and on which their losses depend. How does this inherited structure influence subsequent learning? We examine how the updates are distributed across the spectrum, investigate  the gradient contributions underlying this distribution, and test whether restricting updates to the bulk protects earlier tasks.

\subsection{Later displacement concentrates in the existing upper edge}
\label{sec:traffic}

To examine where these updates land, we read each stage's displacement in the fixed singular basis of its starting checkpoint $W_{i-1}$, i.e., the directions present before the new task is learned. We split its input and output directions into the upper edge ($V_E,U_E$) and the remaining directions ($V_R,U_R$), including the null directions on the longer side of a rectangular matrix. The stage's displacement then splits into
\begin{equation}
  \dW_i=\underbrace{U_E\Delta_{EE}V_E^\top}_{\text{within the upper edge}}
  +\underbrace{U_E\Delta_{ER}V_R^\top+U_R\Delta_{RE}V_E^\top}_{\text{between edge and rest}}
  +\underbrace{U_R\Delta_{RR}V_R^\top}_{\text{within the rest}},
  \qquad \Delta_{XY}=U_X^\top\dW_i V_Y.
  \label{eq:edge-decomposition}
\end{equation}
In the figures, the rest is split further into the bulk, the lower edge and the null directions (\cref{sec:edge}), which divides the displacement into blocks by their input and output directions. For each block, we measure its share of the total displacement energy and, since blocks contain different numbers of entries, compare its mean squared displacement per entry with that of the whole matrix, so that a ratio above one marks a block where displacement concentrates.

\begin{figure}[h]
\centering
 \includegraphics{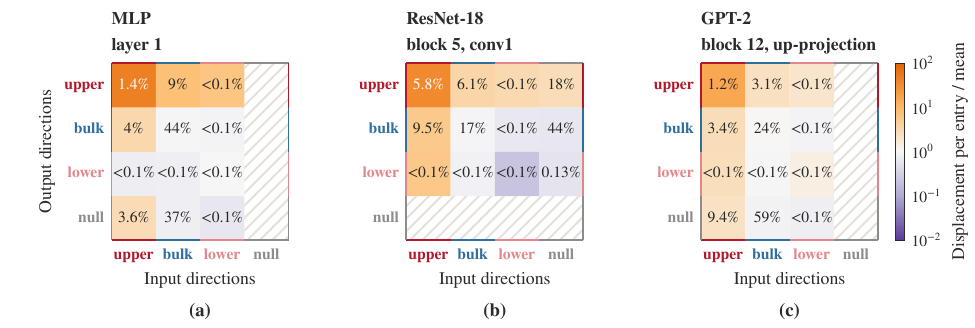}
\caption{\textbf{The next task's displacement concentrates in the existing upper edge.} Panels show displacement during T2 in the singular basis of the starting checkpoint, for \textbf{(a)}~the MLP's expanding hidden matrix (tall), \textbf{(b)}~the first convolution of ResNet-18's residual block 5 (wide) and \textbf{(c)}~the feed-forward up-projection of GPT-2's last block (tall), same layers explored in \cref{fig:stratigraphy}. Blocks pair output directions (rows) with input directions (columns) among the upper edge, bulk, lower edge and null directions (\cref{sec:edge}), and a hatched block is empty. Color is the mean squared displacement per entry relative to the matrix mean, on a log scale with ×1 at the average, and numbers give each block's share of the displacement energy in percent. Every stage of the inspected layers, and all hidden layers pooled, are in \cref{app:displacement-results}.}
\label{fig:traffic}
\end{figure}

Across the three settings, the upper edge receives far more displacement energy than its share of matrix entries (\cref{fig:traffic}; \cref{app:displacement-results,app:displacement-depth}).
In the ResNet-18 layer shown in \cref{fig:traffic}, blocks involving upper-edge directions contain 11\% of the entries and receive 39\% of the displacement energy, while the lower edge and the null directions fall below the matrix average per entry except in blocks that also involve upper-edge directions.
Displacement per entry is strongest in $\Delta_{EE}$ at every task boundary in all three settings, at least twice any other block. Within it, displacement falls both on and off the diagonal (\cref{app:displacement-depth}).
In the reference singular basis, diagonal entries change the coefficients of the inherited upper-edge modes, while off-diagonal entries couple different upper-edge input and output directions. We detail the corresponding standard first-order SVD identities \citep{papadopoulo2000estimating} in \cref{app:singular_dynamics}.

The upper-edge subspaces spanned by $U_E$ and $V_E$ remain  largely aligned across a stage (\cref{sec:stratigraphy}). In \cref{eq:edge-decomposition}, the cross terms couple these subspaces to the remaining directions, while $\Delta_{EE}$ modifies the mapping between the inherited input and output edge directions.
Subsequent learning can therefore alter the mapping on which earlier tasks depend, even when the underlying subspaces largely persist.

\subsection{The incoming gradient favors the inherited edge}
\label{sec:ports}

\begin{figure}[h]
\centering
\includegraphics[width=0.96\linewidth]{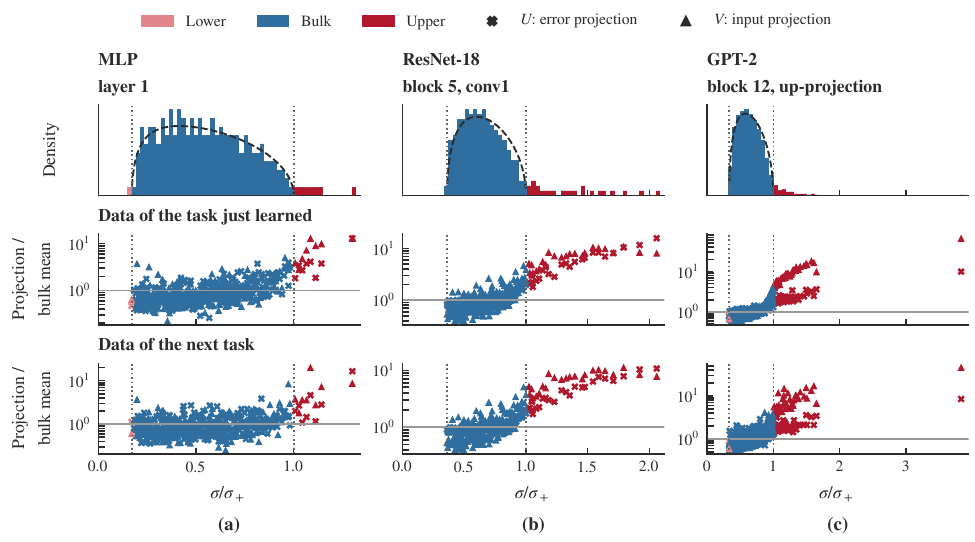}
\setlength{\abovecaptionskip}{5pt}
\caption{\textbf{The incoming task's inputs favor inherited upper-edge directions before training.} Panels show \textbf{(a)}~the MLP, \textbf{(b)}~ResNet-18 and \textbf{(c)}~GPT-2 at the layers of \cref{fig:traffic}, at the end of T1 before training on T2. Rows show the spectrum with the fitted MP bulk (top) and each mode's error projection $\mathbb{E}[(u_k^\top\delta)^2]$ ($U$) and input projection $\mathbb{E}[(v_k^\top h)^2]$ ($V$), on the data of the task just learned (middle) and of the next task (bottom). Each projection is divided by its mean over the layer's bulk modes on the same task data, so values above one, the horizontal line, mark a stronger projection than onto an average bulk direction. Dotted lines mark the MP boundaries. Measurement details are in \cref{app:loadings}.}
\label{fig:ports}

\end{figure}

The displacement of a stage sums the optimizer's updates, which are built from gradients that sum outer products $\delta h^\top$ (\cref{eq:gradient-coefficients}). To see why it lands in the inherited upper edge, we examine the two factors of each contribution, the error projection $u_k^\top\delta$ and the input projection $v_l^\top h$. Since large positive and negative projections can average to zero, we measure their mean squared projections $\mathbb{E}[(u_k^\top\delta)^2]$ and $\mathbb{E}[(v_l^\top h)^2]$ over examples and the positions where the layer is applied. By the Cauchy--Schwarz inequality, the product of these two mean squares bounds the squared gradient entry, up
to a common layer-dependent factor.\footnote{For a layer applied at $A$ positions per example, this factor is $A^2$, since the gradient sums contributions over positions while the projections are averaged over them. The factor cancels when comparing spectral regions within the same layer (\cref{app:loadings}).} Strong error and input projections therefore allow for larger gradient entries, provided that their contributions do not cancel when aggregated. At the end of each task, we compute both on the data of the task just learned and of the next task, with the same fixed weights and singular basis, and divide each by its average over the bulk modes of the same layer and data source (\cref{app:loadings}). On the task just learned, both project strongly onto the upper edge (\cref{fig:ports}, middle row). Learning built the edge from these same factors, since each gradient of that task took its input side from the task's inputs and its output side from its errors. This fits \citet{staats2025small}, who find that the input-side singular vectors of outlying singular values align with principal directions of the input activations.

On the data of T2, on which the network has not yet been trained (\cref{fig:ports}, bottom row), the inputs of the three example layers project onto the upper edge 6.3 to 7.3 times more strongly than onto the bulk, much as they do on T1's data, while the errors favor it less.

In every network, the inputs and the errors of the upcoming task each favor the upper edge in at least 90\% of hidden matrices, counted at every task boundary and in every seed (\cref{tab:incoming,app:projection-results}).

\begin{wrapfigure}{r}{0.35\linewidth}
\centering
\vspace{-14pt}
\includegraphics[width=\linewidth]{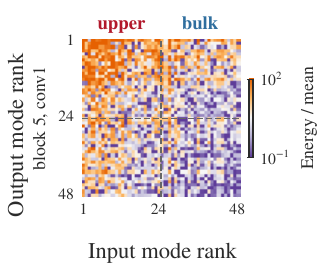}
\raggedright
\setlength{\abovecaptionskip}{4pt}
\caption{\textbf{The incoming gradient is largest where upper-edge inputs meet upper-edge errors.}
Squared entries of $U^\top\widehat G V$ for T2 at the end of T1 (ResNet-18, block 5, conv1), relative to their mean, with the upper edge dashed (\cref{app:gradient-depth}).}
\label{fig:gradient-detail}
\vspace{-20pt}
\end{wrapfigure}

The inputs project more weakly onto the lower edge than onto the bulk in the MLP and ResNet-18, but far more strongly in GPT-2's feed-forward up-projections, whose few lower-edge modes align with the incoming activations, as \citet{staats2025small} find for this architecture type (\cref{app:projection-results}). Inputs and errors reach a layer from opposite directions. The shared layers below produce the new task's inputs, which favor the same directions as the inputs that built the edge. Errors start from the new task's loss and pass backward through the trained layers above. They reflect the new targets and favor the upper edge less strongly than the inputs do.

Because the squared projections only bound the gradient, we also estimate the next task's loss gradient $\widehat G$ before any update, and read it in the inherited singular basis, $U^\top\widehat G V$.
In the ResNet-18 zoom (\cref{fig:gradient-detail}), its squared entries are largest where upper-edge inputs meet upper-edge errors, where both projections of \cref{fig:ports} are large.
In every inspected layer and at every task boundary, blocks involving upper-edge directions receive 2.2 to 8.9 times the mean gradient energy per entry of the whole matrix, and they are enriched in nearly every hidden matrix (\cref{app:gradient-depth}).
Consistent with the inputs favoring the upper edge more than the errors do, the gradient also runs more strongly from upper-edge inputs into the bulk than the other way, by 3.2 to 7.9 times per entry across the three settings (\cref{fig:gradient-detail,tab:incoming}; \cref{app:loadings}).
The displacement concentrates in the inherited upper edge in all three networks (\cref{fig:traffic}). Averaged over layers, seeds and task boundaries, it also favors the block from upper-edge inputs to bulk outputs in each network (\cref{app:displacement-results}).
The gradient thus points at the upper edge from the first step.

\subsection{Protecting the edge reduces forgetting and slows the erosion of pretraining}
\label{sec:freeze}

Undoing a later stage's edge-involving changes partly restores earlier tasks (\cref{sec:fossil}). We now test whether keeping those changes out during training reduces forgetting more than protecting as many bulk directions.
Before learning a new task, we add the upper- and lower-edge directions of its starting checkpoint to those protected for earlier tasks, hold the projectors $\Pout$ and $\Pin$ onto all protected output and input directions fixed during the stage, and remove the part of each proposed update $\dW$ involving them,
\begin{equation}
  \dW \leftarrow (I-\Pout)\,\dW\,(I-\Pin).
  \label{eq:edge-projection}
\end{equation}

Learning can then change only connections between unprotected input and output directions, including the null directions, which the mapping defined by the starting checkpoint leaves unused.
We project both the gradient and the optimizer's update, so momentum, adaptive rescaling and weight decay cannot reintroduce protected directions.
As a control, bulk protection removes the bulk-involving part instead. In each layer, it protects randomly chosen bulk modes, as many as edge protection protects in the upper and lower edge combined (\cref{app:protection}).

\begin{figure}[H]
\centering
\includegraphics[width=0.9\linewidth]{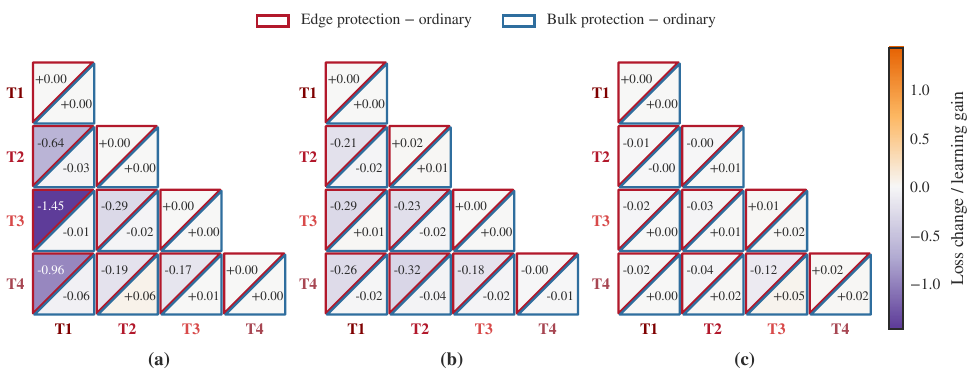}
\setlength{\abovecaptionskip}{5pt}
\caption{\textbf{Edge protection lowers the loss of earlier tasks at little cost to new ones.} Panels show \textbf{(a)}~the MLP, \textbf{(b)}~ResNet-18 and \textbf{(c)}~GPT-2. Rows are completed stages and columns evaluated tasks, so the diagonal holds the task learned in that stage and the cells below it earlier tasks. In each cell, the two triangles compare the validation loss under edge and under bulk protection with ordinary training, in units of the task's learning gain under ordinary training for the same seed, averaged over five seeds, so negative values mean a lower loss than ordinary training. The first stage is shared by all three configurations. Full loss trajectories are in \cref{app:trajectory-results}.}
\label{fig:freeze}
\end{figure}

Edge protection reduces final forgetting more than bulk protection in all three settings at a small cost to new tasks (\cref{fig:freeze}; \cref{fig:freeze-curves} in \cref{app:trajectory-results}; \cref{tab:protection} in \cref{app:protection}).
In ResNet-18, edge protection nearly halves final forgetting on T1--T3, from 54\% of the ordinary learning gain to 28\%, compared with 51\% under bulk protection.
On T2--T4, edge protection reaches nearly the same final validation loss as ordinary training, finishing higher by just 0.5\% of the ordinary learning gain on average.
In the MLP and GPT-2, it lowers forgetting by about a third and a tenth, at a cost below 1\% of the learning gain (\cref{tab:protection}).
Bulk protection yields about as much forgetting as ordinary training in the MLP and ResNet-18.
In the MLP and ResNet-18, where little of the displacement reaches the lower edge (\cref{sec:traffic}), protecting only the upper edge reduces forgetting about as much as protecting both edges, at a similar cost to new tasks (\cref{app:upper-protection}).

Pretraining is GPT-2's earliest task, and held-out WebText loss measures how much of it survives fine-tuning.
Protecting the edge from the pretrained checkpoint onward cuts the WebText loss rise after four classification tasks from 0.31 to 0.24 nats per token (\cref{fig:finetune}).
Bulk protection increases the loss rise relative to ordinary fine-tuning after every task in every seed.

\begin{minipage}[c]{0.48\linewidth}
    Adding the lower edge to upper-edge protection reduces the WebText loss rise from the second task onward in every seed (\cref{app:upper-protection}).
Mean accuracy on the new tasks varies by less than one percentage point across configurations, and earlier classification tasks are forgotten about equally under ordinary fine-tuning, bulk protection and edge protection (\cref{fig:gpt2-trajectories} in \cref{app:trajectory-results}).

\begin{takeaway}
Later learning lands in the same edge that earlier tasks depend on.
A new task's displacement concentrates in the existing upper edge, modifying the mapping on which earlier tasks depend while largely preserving the edge subspaces.
Its gradient is already concentrated there before training, consistent with the strong projection of its inputs and errors onto the upper edge.
Protecting the edge reduces forgetting more than protecting the same number of bulk directions in each matrix, at a small cost to new tasks, and in a pretrained model it slows the erosion of pretraining by about a fifth.
\end{takeaway}
\end{minipage}
\hfill
\begin{minipage}[c]{0.48\linewidth}
    \centering
    \includegraphics[width=0.8\linewidth]{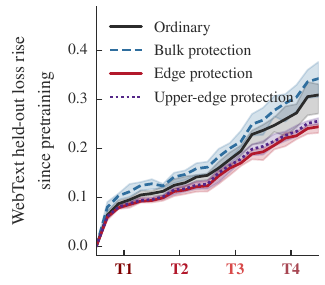}
    \captionof{figure}{\textbf{Protecting the edge slows the erosion of pretraining, and protecting the bulk speeds it up.} WebText held-out loss rise of pretrained GPT-2 during fine-tuning on four text classification tasks (T1--T4), without protection or with bulk, upper-edge or full-edge protection. Edge protection reduces the final rise by about a fifth, while bulk protection increases it by about a tenth. Lines show the mean over five seeds and bands their range.}
    \label{fig:finetune}
\end{minipage}

%% file: sections/05_related.tex
\section{Related work}
\label{sec:related}

\paragraph{Protecting earlier tasks.}
Continual-learning methods control which changes the network can make, either by penalizing changes to parameters weighted by their diagonal Fisher information \citep{kirkpatrick2017overcoming}, a Kronecker-factored estimate of their curvature \citep{ritter2018online}, or their accumulated gradient importance \citep{zenke2017synaptic}, or by keeping updates orthogonal to  the subspaces that earlier tasks' activations occupy, selected by an energy threshold \citep{saha2021gradient}  or by their small singular values \citep{pham2026learning}.
These methods use gradients or activations from earlier tasks. Our edge protection (\cref{sec:freeze}) instead identifies the protected directions from the spectrum of the weights alone.
Other work locates forgetting mainly in deeper representations and recovers earlier performance by resetting those layers \citep{ramasesh2021anatomy}, while we examine forgetting within individual weight matrices.

\paragraph{Spectral fine-tuning.}
Recent work has explored the use of spectral information to preserve previously acquired knowledge during fine-tuning pretrained models.
Methods in the literature keep fine-tuning updates away from the dominant singular directions, either by projecting them onto the complement of the leading singular subspace on both sides \citep{xiong2026oplora,zhang2026spectral} or by adapting only the minor singular components \citep{wang2025milora}, and continual variants keep each task's  low-rank  update orthogonal to those of earlier tasks \citep{wang2023olora}  or protect the leading directions found at each task \citep{nayak2026sculpting}.
Analyses of fine-tuning support their premise, showing that the forgetting caused by low-rank  adaptation sits in new dominant singular directions  that accumulate over a sequence of tasks \citep{shuttleworth2025lora}, and that supervised fine-tuning rotates the leading singular subspaces \citep{zhu2026path}.
These methods choose the protected directions by a rank or energy threshold, while our edge protection uses the same two-sided projection as OPLoRA \citep{xiong2026oplora} and REVIVE \citep{zhang2026spectral}, with the protected directions given by the fitted MP  bulk at both ends of the spectrum and accumulated across tasks.
Our analyses  show why these directions matter, since they retain earlier learning  and the incoming task's inputs and errors already  project strongly onto them.

\paragraph{Random matrix theory.}
Random matrix theory describes the spectra of large random matrices, from Wigner's semicircle law for symmetric matrices \citep{wigner1955characteristic} to the Marchenko--Pastur law for the squared singular values of rectangular ones \citep{marvcenko1967distribution}, and so provides a reference for identifying learned structure in the weights.
Trained weight spectra can retain much of the random-matrix bulk, with learning showing up as outlying singular values and heavy tails \citep{martin2021implicit} and as departures in the singular vectors of the largest singular values \citep{thamm2022random}.
In pretrained transformers, the spectrum departs from the bulk at both ends, the departing directions overlap with eigenvectors of the input-activation covariance, and removing them raises perplexity far more than removing bulk directions \citep{staats2025small}. Selective removal of the smallest components of late MLP matrices can recover rarely seen facts \citep{sharma2024truth}, while slightly worsening language-model perplexity. \citet{staats2025small} further show that the effect of removing small singular components depends on whether removal precedes or follows fine-tuning.
These analyses examine trained weights.
We extend the comparison of edge and bulk to the weight change made by each task, and follow the edge across a sequence of tasks to link it to both the learning and the forgetting of each one.

%% file: sections/06_conclusion.tex
\section{Conclusion}
\label{sec:conclusion}

In this work, we studied catastrophic forgetting from a spectral perspective by examining the model's weights, using random matrix theory as a reference to identify spectral structure emerging during training. 
Across control, vision and language tasks, we found that changes involving the spectral edge, representing a minority of the networks' modes,
contribute disproportionately to both the learning of new tasks and the forgetting of earlier ones. Moreover, the upper edge largely persists across tasks, retaining many previously formed modes as new ones emerge.
A new task's inputs and errors strongly project onto this edge even before training begins, consistent with the concentration of subsequent weight changes that alter the inherited mapping while largely preserving its underlying subspaces.
Protecting the edge, which relies solely on the weights, reduces forgetting more than protecting an equal number of bulk directions, at a small cost to new learning. In a pretrained language model, it also slows the erosion of pretraining performance during sequential fine-tuning.
Previous and subsequent learning thus draw on the same spectral directions. In this sense, forgetting arises when new learning alters directions that earlier tasks depend on.

Despite protecting the inherited edge, updates through the remaining directions can still alter the representations on which earlier tasks depend. This motivates investigating the complementarity of edge protection with methods that protect activations or other parameters. Furthermore, as the protected spans grow across tasks, fewer input and output directions remain available for new updates, motivating future work on how to manage this protection over longer task sequences.

%% file: appendix/proof.tex
\clearpage
\section{Derivations: Singular value and singular vector dynamics}
\label{app:singular_dynamics}

For completeness, we reproduce the derivation of the standard first-order SVD differentiation identities \citep{papadopoulo2000estimating} in our notation.

\begin{proof}
Let $W(t)=U(t)\Sigma(t)V(t)^\top$ be a differentiable weight trajectory with distinct, positive singular values, and let $k$ and $l$ range over its $p=\min(m,n)$ modes. For brevity, write
\[
J=U^\top\dot W V,
\qquad
J_{kl}=u_k^\top\dot W v_l.
\]
For every singular mode $k$,
\[
Wv_k=\sigma_k u_k,
\qquad
W^\top u_k=\sigma_k v_k.
\]
Differentiating the first identity gives
\begin{equation}
\dot Wv_k+W\dot v_k
=
\dot\sigma_k u_k+\sigma_k\dot u_k.
\label{eq:svd-derivative}
\end{equation}
Since the singular vectors are normalized,
\[
u_k^\top\dot u_k=0,
\qquad
v_k^\top\dot v_k=0.
\]
Multiplying \eqref{eq:svd-derivative} from the left by $u_k^\top$ therefore gives
\[
\dot\sigma_k
=
u_k^\top\dot Wv_k
=
J_{kk}.
\]

Now fix $k\neq l$. Multiplying \eqref{eq:svd-derivative} from the left by $u_l^\top$, and using
\[
u_l^\top u_k=0,
\qquad
u_l^\top W=\sigma_l v_l^\top,
\]
gives
\begin{equation}
\sigma_k u_l^\top\dot u_k
-
\sigma_l v_l^\top\dot v_k
=
J_{lk}.
\label{eq:svd-left-rotation}
\end{equation}
Similarly, differentiating $W^\top u_k=\sigma_k v_k$ and multiplying from the left by $v_l^\top$ gives
\begin{equation}
-\sigma_l u_l^\top\dot u_k
+
\sigma_k v_l^\top\dot v_k
=
J_{kl}.
\label{eq:svd-right-rotation}
\end{equation}
Equations~\eqref{eq:svd-left-rotation} and
\eqref{eq:svd-right-rotation} form the linear system
\[
\begin{pmatrix}
\sigma_k & -\sigma_l\\
-\sigma_l & \sigma_k
\end{pmatrix}
\begin{pmatrix}
u_l^\top\dot u_k\\
v_l^\top\dot v_k
\end{pmatrix}
=
\begin{pmatrix}
J_{lk}\\
J_{kl}
\end{pmatrix}.
\]
Its determinant is $\sigma_k^2-\sigma_l^2$, which is nonzero because the singular values are distinct. Inverting the system yields
\[
u_l^\top\dot u_k
=
\frac{
\sigma_k J_{lk}+\sigma_l J_{kl}
}{
\sigma_k^2-\sigma_l^2
},
\]
and
\[
v_l^\top\dot v_k
=
\frac{
\sigma_l J_{lk}+\sigma_k J_{kl}
}{
\sigma_k^2-\sigma_l^2
}.
\]
Substituting $J_{kl}=u_k^\top\dot Wv_l$ gives the singular-vector derivatives in the original coordinates.
For a rectangular matrix, multiplying \eqref{eq:svd-derivative} by a unit vector $u_\perp$ outside the span of $U$, where $u_\perp^\top W=0$, gives $u_\perp^\top\dot u_k=u_\perp^\top\dot W v_k/\sigma_k$,
so the rotation towards the null directions is proportional to the corresponding update component and inversely proportional to $\sigma_k$.
The input side follows in the same way.

\end{proof}

%% file: appendix/b_setups.tex
\section{Experimental setups}
\label{app:setups}

Each setting consists of four tasks learned in sequence without replay.
For each seed, we train the first task once from a random initialization and continue from the resulting checkpoint with ordinary training, edge protection, upper-edge protection and bulk protection. The configurations share the architecture, data order, optimizer recipe, training budget and evaluation examples, and differ only in the hidden-matrix directions they protect. GPT-2 fine-tuning starts every configuration from the same pretrained weights instead. \Cref{tab:setups} lists the tasks, models and optimizers, and the paragraphs below give what the table leaves out.

\begin{table}[h]
\centering
\small
\setlength{\tabcolsep}{3pt}
\caption{Tasks, models and optimization in the three settings and in GPT-2 fine-tuning. Learning-rate schedules restart at every task.}
\label{tab:setups}
\begin{tabular}{lllll}
\toprule
 & MLP & ResNet-18 & GPT-2 & GPT-2 fine-tuning \\
\midrule
Tasks & MimicGen D1 & CIFAR-100 & text domains & text classification \\
 & 4 manipulation skills & $4\times25$ classes & 4 domains & 4 datasets \\
Loss & mean squared error & cross-entropy & next-token cross-entropy & next-token cross-entropy \\
\midrule
Optimizer & SGD & SGD, momentum 0.9 & AdamW & AdamW \\
Learning rate & 0.1, constant & 0.1, cosine & $6\times10^{-4}$, cosine & $2\times10^{-5}$, linear decay \\
Warmup & none & 1 epoch & 0.1 epoch (T1) & 10\% of each task \\
 & & & 200 steps (T2--T4) & \\
Weight decay & 0 & 0 & 0.1 & 0 \\
Batch size & 256 & 128 & 480 & 32 \\
Epochs per task & 100 & 50 & 1 & 3 \\
Precision & float32 & bfloat16 & bfloat16 & bfloat16 \\
\bottomrule
\end{tabular}
\end{table}

\paragraph{MLP.}
The MLP learns square, threading, coffee and three-piece assembly, in that order, from the D1 demonstrations of MimicGen \citep{mandlekar2023mimicgen}. It maps 66 state features of the robot arm and the objects to a 7-dimensional action. We hold out a fifth of the episodes of each task for evaluation, sample 121,856 training transitions per task and standardize the features with their training statistics. The network has a bias-free input layer of width 256 and two hidden matrices, which expand to width 512 and contract back to 256, with GELU activations and a linear action head for each task.

\paragraph{ResNet-18.}
Each seed draws its own random class order. Each task holds out a stratified tenth of its training images for validation, and training images are randomly cropped and flipped. ResNet-18 \citep{he2016deep} has a $3\times3$, stride-one stem without max-pooling, bias-free convolutions and GroupNorm with 32 groups in place of BatchNorm, so that each image is normalized on its own. Each task has its own 25-class head.

\paragraph{GPT-2.}
GPT-2 learns English Wikipedia text from M2D2's CULT domain \citep{reid2022m2d2}, mathematical web text from OpenWebMath \citep{paster2024openwebmath}, Python code from CodeParrot \citep{codeparrot2021clean} and German web text from FineWeb2 \citep{penedo2025fineweb2}, in that order, in sequences of 1,024 tokens from the GPT-2 tokenizer. For the three later sources, documents are separated by an end-of-text token and packed into consecutive sequences, and the validation sequences precede the training sequences in each stream, so the two splits share at most the one document that crosses their boundary. Each evaluation uses the same 2,000 validation sequences per task. The model is GPT-2 small \citep{radford2019language}, 12 blocks of width 768 with 12 attention heads, trained from scratch without dropout, with a tied token embedding and output matrix over a vocabulary of 50,304 tokens. AdamW uses $\beta_2=0.95$ and decays the embeddings and weight matrices but not the normalization parameters or biases.

\paragraph{GPT-2 fine-tuning.}
\label{app:gpt2-finetune}
We fine-tune the pretrained GPT-2 small \citep{radford2019language} on DBpedia, Amazon, Yahoo Answers and AG News, the first task order of O-LoRA \citep{wang2023olora}, with 1,000 training examples per class. Each example is a prompt that asks for the topic or sentiment of a paragraph, lists the options and ends with \texttt{Answer:}, and the loss covers the answer label and the end-of-text token. Accuracy requires that the most likely next token match at every scored position under teacher forcing, including the end-of-text token, on 2,000 held-out examples per task. To measure what the model keeps from pretraining, we evaluate the next-token loss on 2,000 documents from GPT-2's WebText validation set. Edge and bulk protection start from the pretrained checkpoint, so the first fine-tuning task is already protected, and add the directions of each later starting checkpoint. Gradients are clipped at norm 1.0.

\paragraph{Parameter scope.}
Spectral analysis and interventions act on the hidden weight matrices, which map one learned representation to the next: the two hidden matrices of the MLP, the 19 convolutions of ResNet-18 after its stem, and the 72 attention and feed-forward matrices of GPT-2. From the second task on, only these matrices and the task heads learn. The input layer (the first layer of the MLP, the stem of ResNet-18, and the token and position embeddings of GPT-2 together with the output layer tied to the token embedding) stay fixed, along with normalization scales and all biases outside the task heads. In fine-tuning, GPT-2 keeps all of these at their pretrained values throughout.

\paragraph{Inspected layers.}
\label{app:inspected}
We complement the pooled figures with results for layers at different depths and aspect ratios. These are the two hidden matrices of the MLP, which expand and contract, the first convolutions of residual blocks 3 and 5 and the shortcut projection of block 7 of ResNet-18, and the attention key projection of block 1, the feed-forward down-projection of block 7 and the feed-forward up-projection of block 12 of GPT-2. The main text shows one of them per network, the expanding matrix of the MLP, the convolution of block 5 of ResNet-18 and the up-projection of block 12 of GPT-2.

%% file: appendix/a_spectral_edge.tex
\section{Measurement details}
Pooled spectral quantities are computed within each hidden matrix and averaged over the matrix and seed pairs in which they are defined, with each pair weighing equally. Counts of edge modes are summed over matrices before averaging over seeds.

\label{app:spectral-edge}

\subsection{Spectral fitting and matrix views}
\label{app:mp-fit}
\label{app:matrix-views}
Every matrix is read in output-by-input coordinates.
A convolution of kernel size $k_h\times k_w$ with $c_{\rm in}$ input and $c_{\rm out}$ output channels is read as a $c_{\rm out}\times(k_hk_wc_{\rm in})$ matrix acting on an input patch.
Attention projections are read as their corresponding linear maps, with head dimensions combined where needed.
The same views are used for spectral measurement and intervention.

For a matrix of shape $m\times n$, let $\gamma=\min(m,n)/\max(m,n)$ be the aspect ratio.
For random matrices with independent, centered entries of variance $s/\max(m,n)$, the empirical distribution of squared singular values converges to the MP law with scale $s$ as both dimensions grow at fixed $\gamma$.
We fit each matrix separately at each checkpoint, following \citet{staats2025small} in using the median-based noise estimate of \citet{gavish2014optimal}.
Writing $q_{1/2}(\gamma)$ for the median of the unit-scale MP law, the fitted scale and singular-value cutoffs are
\begin{equation}
 \widehat{s}
 = \frac{\operatorname{median}_{k}\sigma_k^2}{q_{1/2}(\gamma)},
 \qquad
 \sigma_\pm=\sqrt{\widehat{s}}(1\pm\sqrt{\gamma}).
\end{equation}
The fit uses all $p=\min(m,n)$ singular values, without iteratively trimming outliers or choosing a retained rank.

\begin{figure}[H]
\centering
\includegraphics[width=\linewidth]{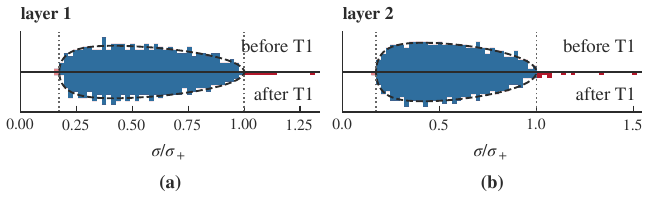}
\includegraphics[width=\linewidth]{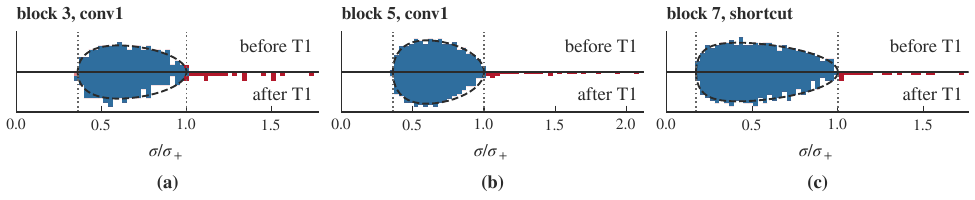}
\includegraphics[width=\linewidth]{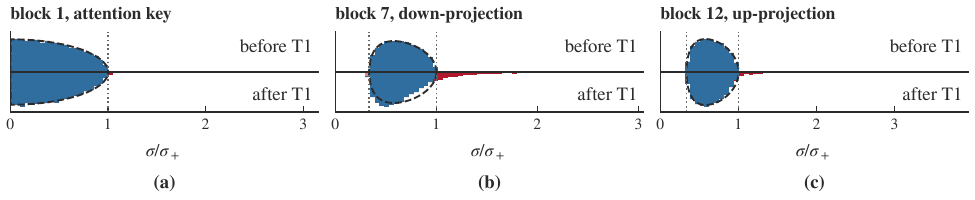}
\caption{\textbf{The fitted MP law describes the spectra at initialization.} From top to bottom, the inspected layers of the MLP, ResNet-18 and GPT-2, with singular values before the first task above the axis and after it below, relative to the upper MP boundary. Dotted lines mark the fitted MP boundaries.}
\label{fig:init-spectra}
\end{figure}

At initialization, fewer than 0.2\% of hidden modes per network lie outside the fitted bulk, averaged over seeds (\cref{fig:init-spectra}).
Spectrum figures plot $\sigma/\sigma_+$, with cutoffs at $\sigma_-/\sigma_+$ and $1$.
Spectral measurements use float32 even when training uses bfloat16 (\cref{tab:setups}).

\subsection{Persistence and mode tracking}
\label{app:continuity}
\label{app:mode-tracking}
Persistence uses a separate MP fit at the start and end of each stage.
For each region, we compute the persistence of \cref{eq:persistence} and its chance level for each matrix and seed in which the region is nonempty at both ends of the stage, and average both equally over those pairs.
The chance level of a region is its share of the modes, $|\mathcal{R}'|/\min(m,n)$, the persistence that a unit direction drawn uniformly at random within the span of all $p=\min(m,n)$ singular vectors on that side at the subsequent checkpoint would reach. A direction drawn uniformly in the whole input or output space would reach only $|\mathcal{R}'|/\max(m,n)$ on the longer side, so the share of modes is the higher, stricter chance level.
On each side, the double sum in \cref{eq:persistence} equals the sum of the squared cosines of the $\min(|\mathcal{R}|,|\mathcal{R}'|)$ principal angles between the region's subspaces before and after the stage.
Dividing by $|\mathcal{R}|$ measures the geometric overlap relative to the dimension of the earlier subspace.
A region whose later subspaces lie inside its earlier ones with half their dimension therefore has persistence $1/2$, even though all its principal angles are zero.
Mode shares in \cref{fig:stratigraphy} are the fractions of a layer's modes in each region at the end of the stage, averaged over seeds.

Mode matching compares individual modes rather than regions.
Since the singular vectors have unit length, each rank-one map $u_kv_k^\top$ has unit Frobenius norm, and its overlap with a mode $l$ at the next checkpoint is
\begin{equation}
 \overlap_{kl}
 = \langle u_kv_k^\top,u'_lv_l'^\top\rangle_F
 = (u_k^\top u'_l)(v_k^\top v'_l).
 \label{eq:overlap}
\end{equation}
This score compares the input and output directions independently of their singular values, so a mode can be followed while its singular value changes. The overlap is unchanged when both singular vectors of a mode change sign. Its sign tells whether a matched map keeps or reverses its orientation. The assignment maximizes the signed overlap, and the figures show $\overlap_{kl}$.

For each hidden matrix and seed, we match its full spectrum of $p=\min(m,n)$ modes between consecutive checkpoints.
Hungarian assignment selects a permutation $\pi$ maximizing
\begin{equation}
 \sum_{k=1}^{p}\overlap_{k,\pi(k)}.
 \label{eq:mode-assignment}
\end{equation}
The assignment uses no edge or bulk labels, so it can follow modes across either MP boundary.

The checkpoint chain starts immediately before task 1 and continues through every task endpoint.
At each checkpoint we fit the MP boundaries and classify modes as lower edge, bulk or upper edge.
An edge mode whose matched predecessor was also in the edge inherits its predecessor's entry-task label.
A mode matched from bulk into either edge receives the current task's label; a mode matched from edge into bulk is no longer counted as an edge mode.
Modes already in the edge before task 1 retain a separate initial label until they leave it.
These labels give the colors by task of entry in \cref{fig:esd}.

\subsection{Displacement, removal and protection}
\label{app:traffic-more}
\label{app:removal}
\label{app:protection}
The block view of \cref{sec:traffic} partitions the full displacement matrix using the starting checkpoint's lower-edge, bulk and upper-edge subspaces. For rectangular matrices, the directions outside the compact singular basis form the null region of \cref{sec:edge}. In \cref{eq:edge-decomposition}, these directions complete the output or input basis on the longer side, so each term has the shape of $W$. A block's displacement energy is its squared Frobenius norm, and its number of entries is the product of its input and output dimensions. Within each matrix, a block's enrichment is its share of the displacement energy over its share of the entries, and pooled figures average it over the matrices that have the block.

\label{app:gradient-zoom}
The zooms show entries of the compact $p\times p$ matrices $U^\top\dW V$ (\cref{app:displacement-depth}) and $U^\top\widehat G V$ (\cref{fig:gradient-detail}; \cref{app:gradient-depth}) for the inspected layers, with modes ordered by decreasing singular value at each checkpoint. Each squared entry is divided by the mean squared entry over the full matrix, including the rectangular null directions, and the window shows the upper-edge modes and as many adjacent bulk modes.

The removal in \cref{sec:fossil} uses the edge fitted at the end of the stage, when the task's learned modes are present.
The displacement measurements of \cref{sec:traffic} use the edge fitted at the start, before the task is trained.
In both cases the net displacement is the same, $\dW=W_i-W_{i-1}$, and the reference checkpoint decides which directions form its edge.
The rank-matched comparison removes parts involving equal numbers of edge and bulk modes in each matrix. Both samples use the smaller of the two available mode counts. The cap affects four ResNet-18 matrix endpoints across the five seeds.
The energy control rescales these rank-matched edge-involving and bulk-involving parts of each matrix's displacement to the smaller of their two squared Frobenius norms, so that both removals have the same energy in every matrix.

At the end of each stage, edge protection fits the current weights and adds their lower- and upper-edge input and output directions to the spans protected at earlier stages.
We form an orthonormal basis of each accumulated span and keep it fixed throughout the stage.
Bulk protection accumulates randomly selected bulk directions until its protected input and output ranks match those of edge protection in each matrix, which they do in every hidden matrix, stage and seed of all four experiments.

We apply the two-sided projection both to the raw gradient before the optimizer and to the proposed weight update after it. The second projection removes any protected component introduced by momentum, adaptive rescaling or weight decay. Every applied update $\dW$ therefore satisfies $\dW v_k=0$ and $u_k^\top\dW=0$ for protected input and output directions $v_k$ and $u_k$, up to numerical precision. Other trainable parameters follow the ordinary optimizer. GPT-2 trained from scratch uses weight decay, which shrinks each mode in proportion to its singular value, so it adds to the diagonal of $\Delta_{EE}$ and, under edge protection, no longer acts on protected directions. The MLP, ResNet-18 and GPT-2 fine-tuning use no weight decay and show the same upper-edge concentration and protection benefit.

We report two quantities for each configuration, both in units of the learning gain of \cref{sec:forgetting}. The forgetting of an earlier task is the rise of its loss from the end of its own training stage to the end of the sequence, measured within that configuration, so each configuration is compared with the loss at which it acquired the task. The cost to a new task is the difference between the configuration's loss and that of ordinary training on each of tasks 2--4 at the end of its own stage. For each seed, we divide both by the evaluated task's learning gain under ordinary training, which is positive for every task and seed, and average over tasks and seeds without clipping. A negative forgetting, where a task's loss ends below the loss at which it was acquired, therefore enters the means as recovery, and a negative cost marks a new task learned better than under ordinary training.

\begin{table}[H]
\centering
\small
\caption{Forgetting of earlier tasks and cost to new tasks, in units of the learning gain, as seed mean $\pm$ standard deviation. Forgetting averages tasks 1--3 from the end of their own stage to the end of the sequence, and cost averages tasks 2--4 at the end of their own stage against ordinary training.}
\label{tab:protection}
\begin{tabular}{lccccc}
\toprule
 & \multicolumn{3}{c}{Forgetting} & \multicolumn{2}{c}{Cost to new tasks} \\
\cmidrule(lr){2-4}\cmidrule(lr){5-6}
 & Ordinary & Bulk & Edge & Bulk & Edge \\
\midrule
\input{tables/protection}
\end{tabular}
\end{table}

\subsection{Gradient and projection measurements}
\label{app:loadings}
At the end of each task, we pass a fixed set of 1,024 training examples from that task and from the next through the network without updating its weights.
For each layer we record the input $h$ and the loss gradient $\delta$ at its output.
The input and error projections of mode $k$ are the second moments of $v_k^\top h$ and $u_k^\top\delta$, respectively.
A layer applied $A$ times per example, at input patches or token positions, contributes every application, and since its gradient sums over them, the Cauchy--Schwarz bound of \cref{sec:ports} holds up to the factor $A^2$, which is the same for every entry of a layer and cancels in every ratio we report. Input and error projections are each divided by their average over the bulk modes of that layer, so the plotted values compare directions within the same layer and data source.
Edge-to-bulk ratios average these normalized projections over the upper-edge modes.

The bound also compares the two blocks connecting the upper edge and the bulk.
Write $a_k=\mathbb{E}[(u_k^\top\delta)^2]$ and $b_l=\mathbb{E}[(v_l^\top h)^2]$, and $\bar a_{\mathcal{S}}$, $\bar b_{\mathcal{S}}$ for their means over a set of modes $\mathcal{S}$.
The bound on entry $(k,l)$ is proportional to $a_k b_l$, so its average over a block is the product of the two means, and the block from upper-edge inputs $\mathcal{E}$ to bulk outputs $\mathcal{B}$ compares with the reverse block as
\begin{equation*}
  \frac{\bar a_{\mathcal{B}}\,\bar b_{\mathcal{E}}}{\bar a_{\mathcal{E}}\,\bar b_{\mathcal{B}}}
  =\frac{\bar b_{\mathcal{E}}/\bar b_{\mathcal{B}}}{\bar a_{\mathcal{E}}/\bar a_{\mathcal{B}}},
\end{equation*}
the input ratio over the error ratio within each matrix at each checkpoint.
\Cref{tab:incoming} averages the input ratios, error ratios and measured gradient asymmetries separately over matrices, task boundaries and seeds. The measured gradient favors the block from upper-edge inputs to bulk outputs in all three settings.

At each task boundary, we estimate the incoming task's raw loss gradient at the preceding task's endpoint, without changing the weights. We use the same 1,024 incoming-task training examples as the input and error measurements, without data augmentation, and average the signed gradients before projecting into the checkpoint's singular basis and squaring the coefficients.

%% file: tables/protection.tex
MLP & $1.46 \pm 0.24$ & $1.47 \pm 0.22$ & $1.02 \pm 0.19$ & $0.00 \pm 0.00$ & $0.00 \pm 0.00$ \\
ResNet-18 & $0.54 \pm 0.04$ & $0.51 \pm 0.04$ & $0.28 \pm 0.03$ & $0.00 \pm 0.01$ & $0.01 \pm 0.01$ \\
GPT-2 & $0.65 \pm 0.02$ & $0.66 \pm 0.03$ & $0.59 \pm 0.02$ & $0.02 \pm 0.01$ & $0.01 \pm 0.01$ \\
\bottomrule

%% file: appendix/c_controls.tex
\clearpage
\section{Additional results}
\label{app:controls}
\subsection{Removal at equal energy}
\label{app:energy-results}
\begin{figure}[H]
\centering
\includegraphics[width=0.9\linewidth]{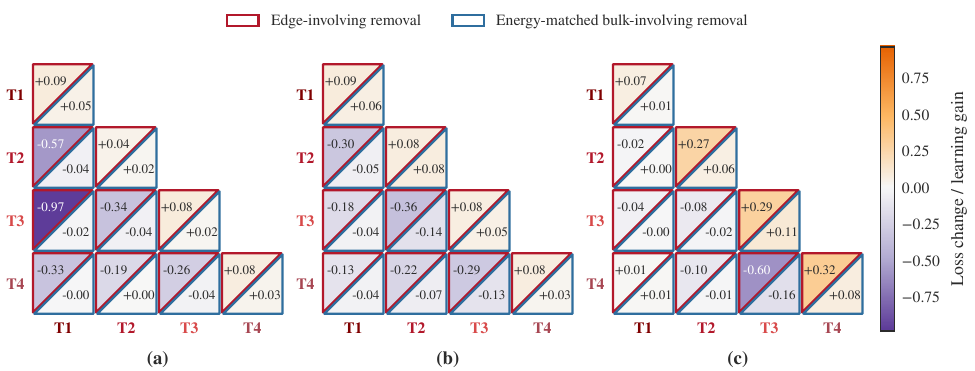}
\caption{\textbf{At equal displacement energy, edge removal leaves lower earlier-task losses than bulk removal.} Panels show \textbf{(a)}~the MLP, \textbf{(b)}~ResNet-18 and \textbf{(c)}~GPT-2. Edge-involving (upper-left) and bulk-involving (lower-right) removals have equal displacement energy within each matrix. Loss changes use the learning-gain normalization of \cref{fig:fossil}; values average five seeds.}
\label{fig:fossil-energy}
\end{figure}

\Cref{fig:fossil-energy} repeats the removal of \cref{fig:fossil} with the rank-matched edge-involving and bulk-involving parts of each matrix's displacement rescaled to the same energy, which separates where a change lies from how large it is (\cref{sec:fossil}). The closest of the 18 earlier-task comparisons is GPT-2's first task after stage 4, at 0.0098 for edge removal and 0.0101 for bulk removal, and edge removal lowers the loss below the intact endpoint in the other 17. Without rescaling, the two comparisons in which edge removal does not lower an earlier task's loss are GPT-2's first task after stages 2 and 4.

\subsection{Persistence across layers}
\label{app:persistence-results}
The upper edge persists far above its chance level when all hidden layers are pooled and in every inspected layer. In ResNet-18 and GPT-2, the pooled lower edge persists less than the upper edge and above its chance level.

\begin{figure}[H]
\centering
\includegraphics[width=\linewidth]{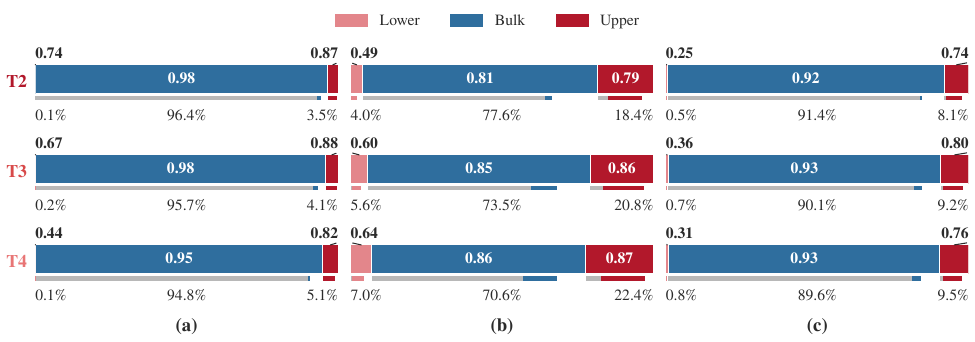}
\caption{\textbf{Persistence pooled over all hidden layers.} Panels show \textbf{(a)}~the MLP, \textbf{(b)}~ResNet-18 and \textbf{(c)}~GPT-2, drawn as in \cref{fig:stratigraphy}. Band widths are averaged over all matrix--seed pairs. For each region, persistence and chance are averaged over the pairs in which the region is nonempty at both checkpoints, with each pair weighted equally.}
\label{fig:persistence-pooled}
\end{figure}

\begin{figure}[H]
\centering
\includegraphics[width=\linewidth]{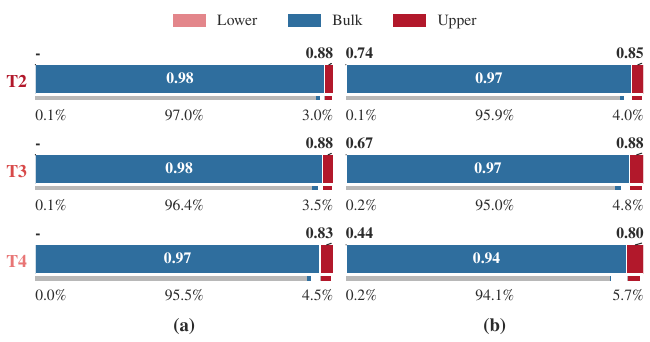}
\includegraphics[width=\linewidth]{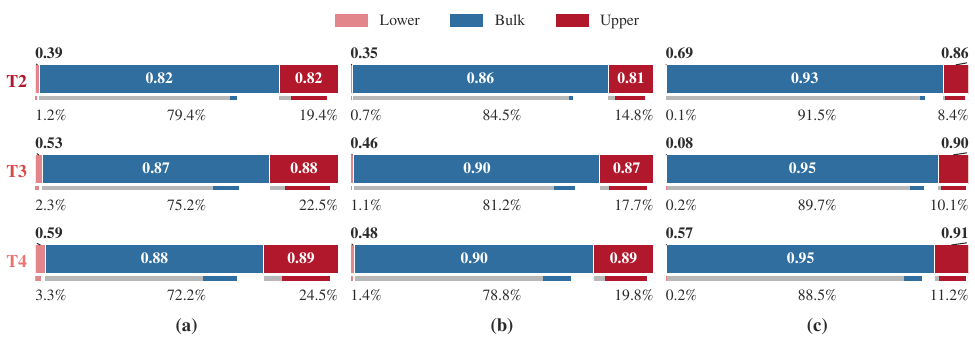}
\includegraphics[width=\linewidth]{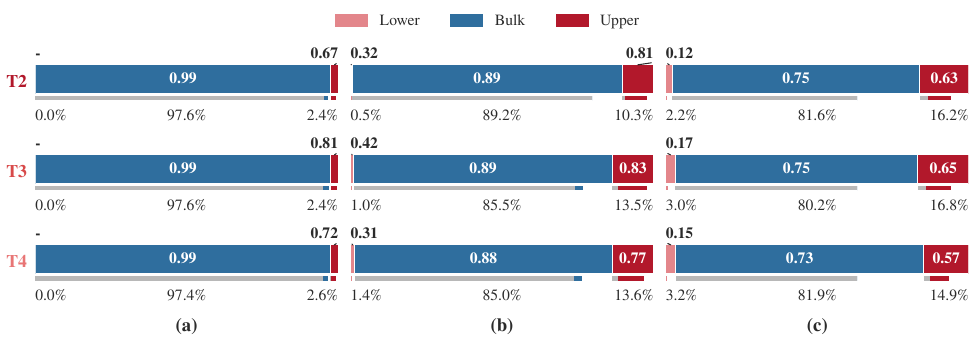}
\caption{\textbf{Persistence in the inspected layers.} From top to bottom, the MLP, ResNet-18 and GPT-2, with columns for the layers of \cref{fig:esd-mlp,fig:esd-resnet,fig:esd-gpt2}, drawn as in \cref{fig:stratigraphy}.}
\label{fig:persistence-layers}
\end{figure}

\clearpage
\subsection{Spectra and mode accumulation}
\label{app:spectra-results}
We show the panels of \cref{fig:esd} for every inspected layer (\cref{sec:esd}). The middle ResNet-18 row is the layer of \cref{fig:esd}. Counted over all hidden matrices and averaged over seeds, the edge keeps 97\% of its first-task modes through the last task in ResNet-18 and the MLP and 81\% in GPT-2. From the end of the first task to the end of the last, it grows by a factor of 2.3 in ResNet-18 and about 1.7 in the MLP and GPT-2.

\begin{figure}[H]
\centering
\includegraphics[width=\linewidth]{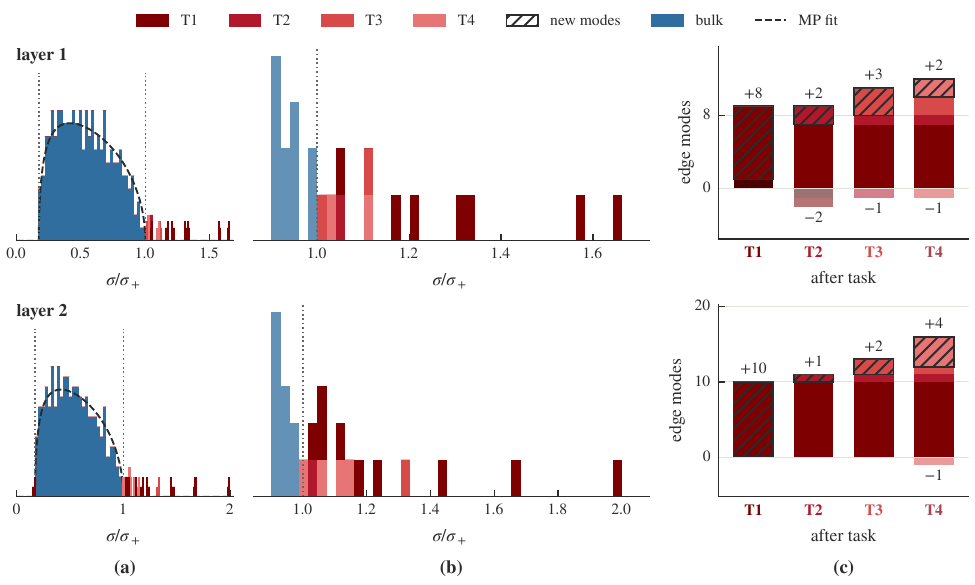}
\caption{\textbf{MLP spectra across depth.} Rows show the expanding and contracting hidden matrices. Columns show \textbf{(a)}~the final spectrum, \textbf{(b)}~its upper edge with modes colored by the task during which they entered it and \textbf{(c)}~the layer's edge modes after each task. Hatching marks newly entered modes; departures are below zero.}
\label{fig:esd-mlp}
\end{figure}

\begin{figure}[H]
\centering
\includegraphics[width=\linewidth]{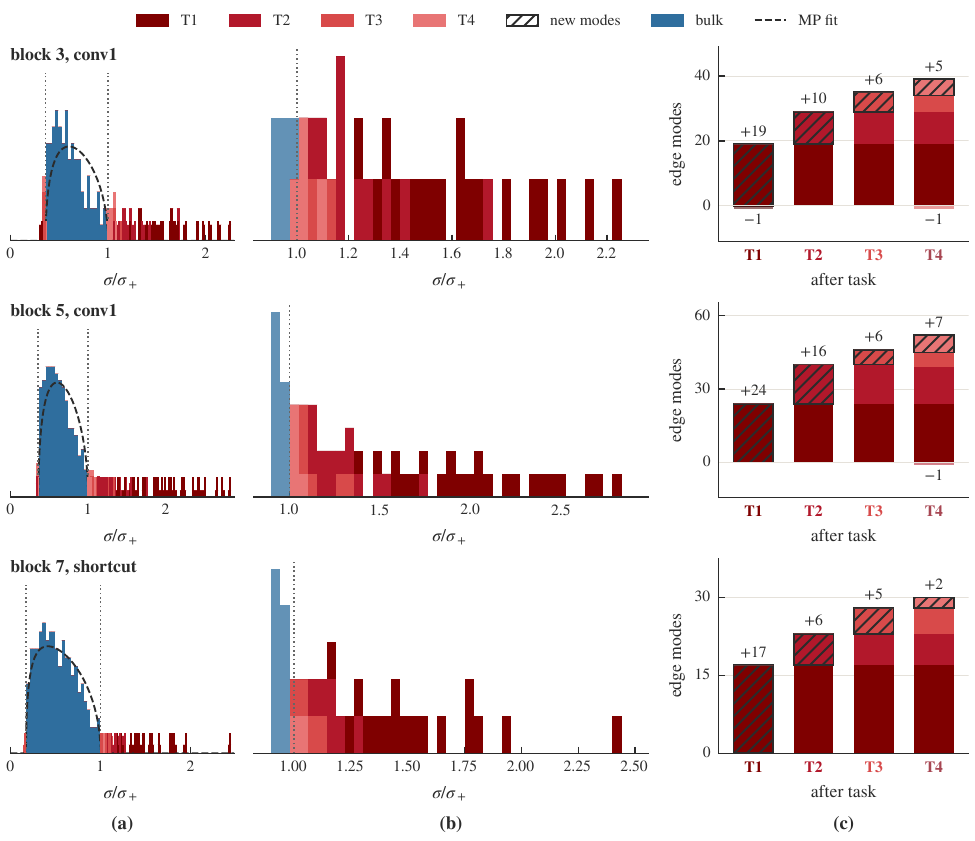}
\caption{\textbf{ResNet-18 spectra across depth.} Rows are the first convolutions of residual blocks 3 and 5 and the shortcut projection of block 7. Panels are drawn as in \cref{fig:esd-mlp}.}
\label{fig:esd-resnet}
\end{figure}

\begin{figure}[H]
\centering
\includegraphics[width=\linewidth]{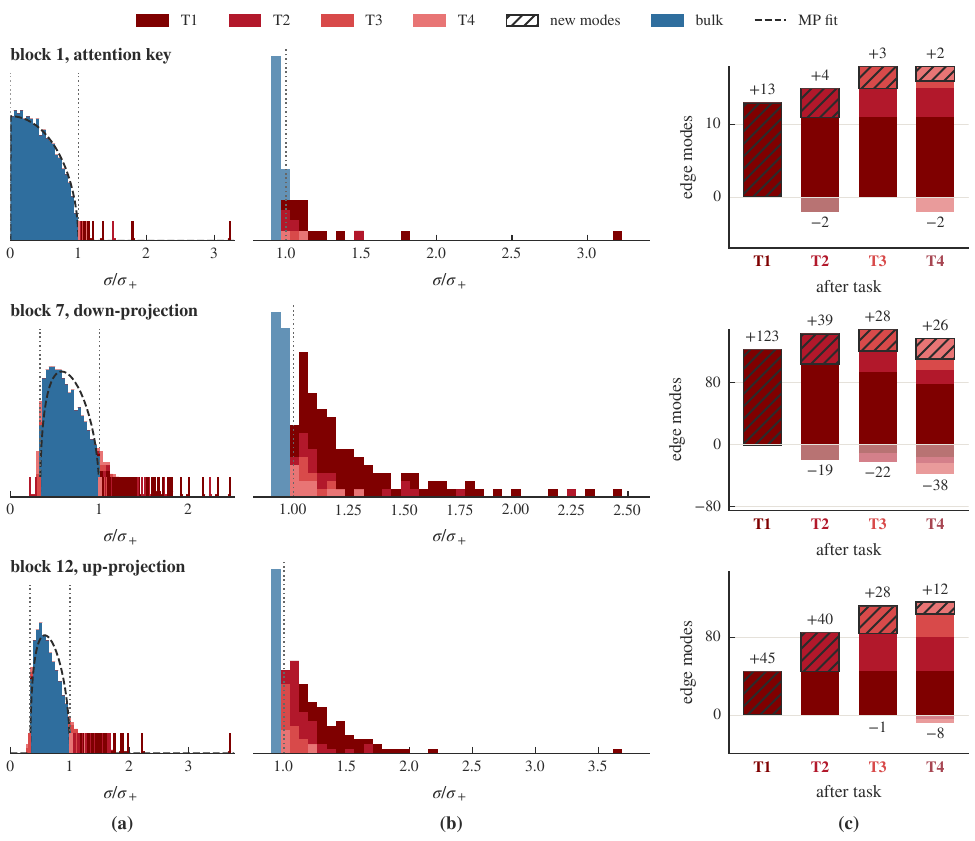}
\caption{\textbf{GPT-2 spectra across depth.} Rows are the attention key projection of block 1, the feed-forward down-projection of block 7 and the feed-forward up-projection of block 12. Panels are drawn as in \cref{fig:esd-mlp}.}
\label{fig:esd-gpt2}
\end{figure}

\clearpage
\subsection{Mode matching across checkpoints}
\label{app:matching}
\begin{figure}[H]
\centering
\includegraphics[width=\linewidth]{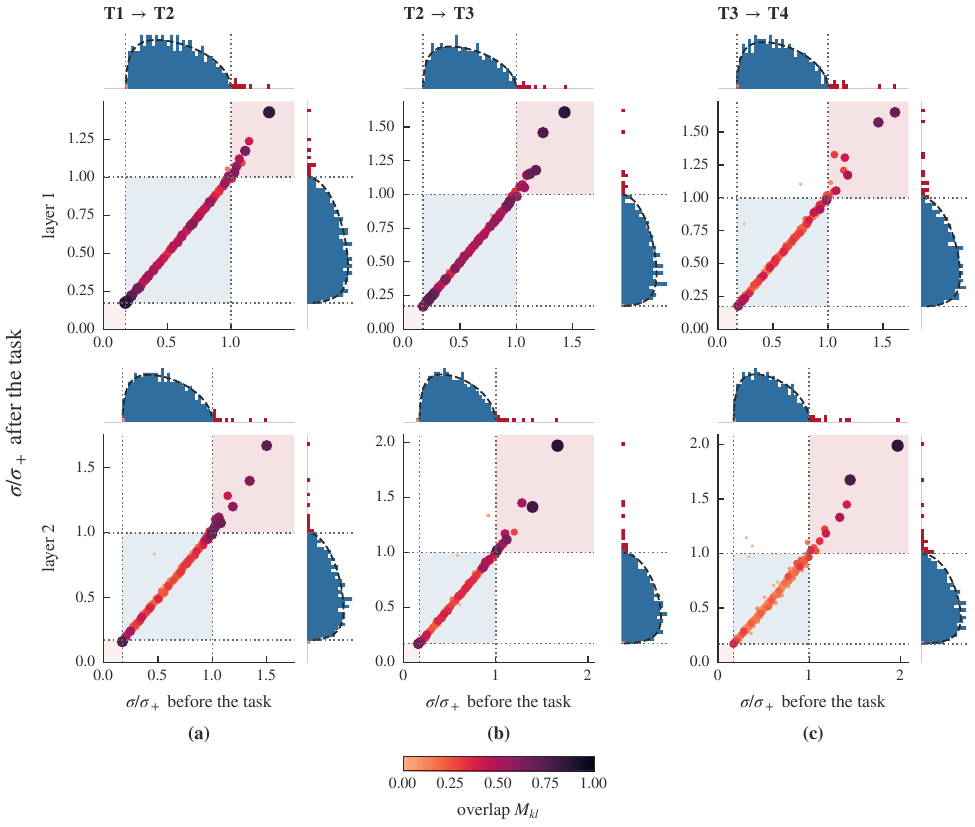}
\caption{\textbf{Matched modes keep similar normalized singular values (MLP).} Each point is a mode at the end of one task and its matched partner at the end of the next, placed at their singular values relative to the corresponding upper MP boundary and colored by their overlap $\overlap_{kl}$. Shaded boxes mark modes that stay in the lower edge, bulk or upper edge, and the spectra before and after the stage run along the axes. Rows are the hidden layers and columns the stages. Dotted lines mark the MP boundaries.}
\label{fig:matching}
\end{figure}

\begin{figure}[H]
\centering
\includegraphics[width=\linewidth]{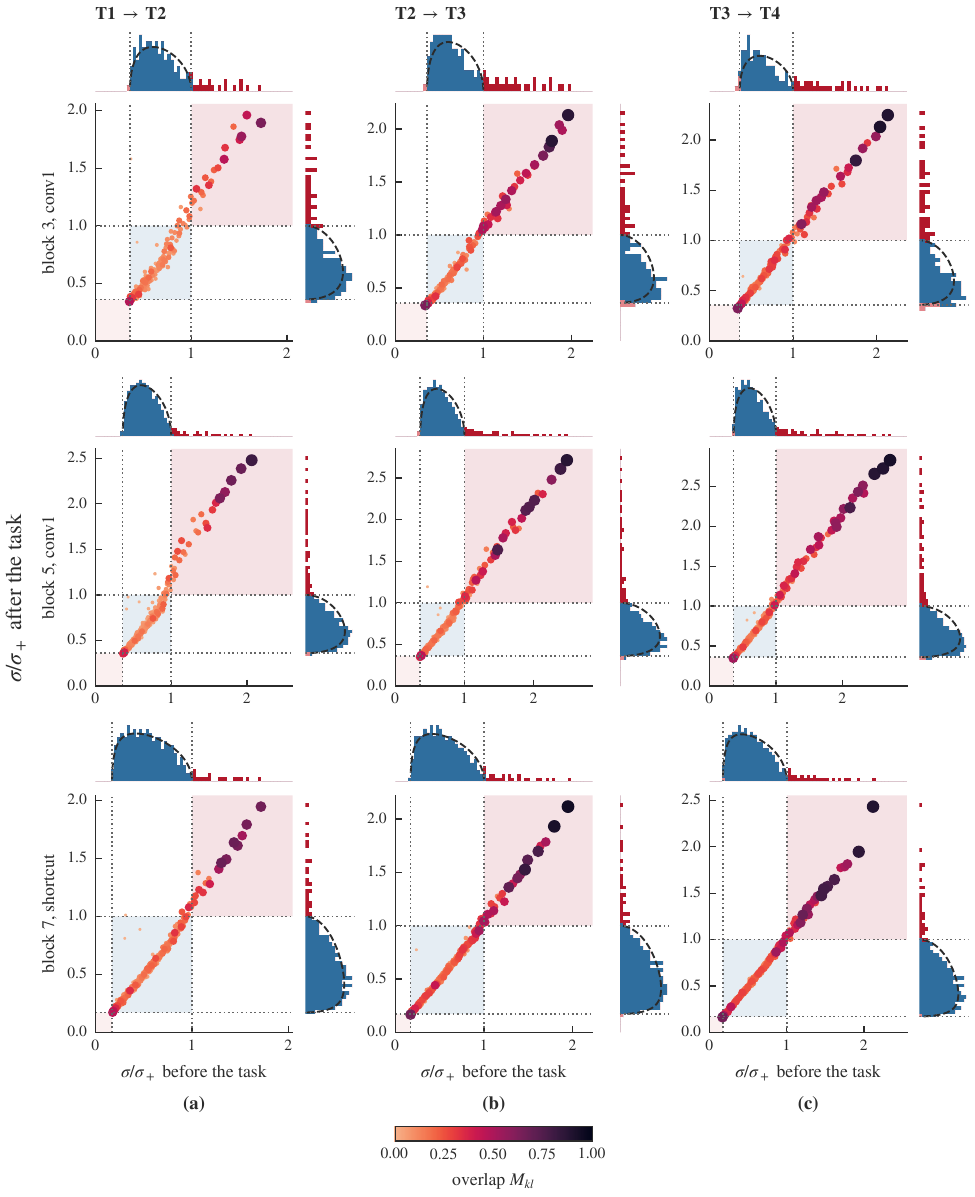}
\caption{\textbf{Matched modes in ResNet-18.} As in \cref{fig:matching}, for the three inspected layers.}
\label{fig:matching-resnet}
\end{figure}

\begin{figure}[H]
\centering
\includegraphics[width=\linewidth]{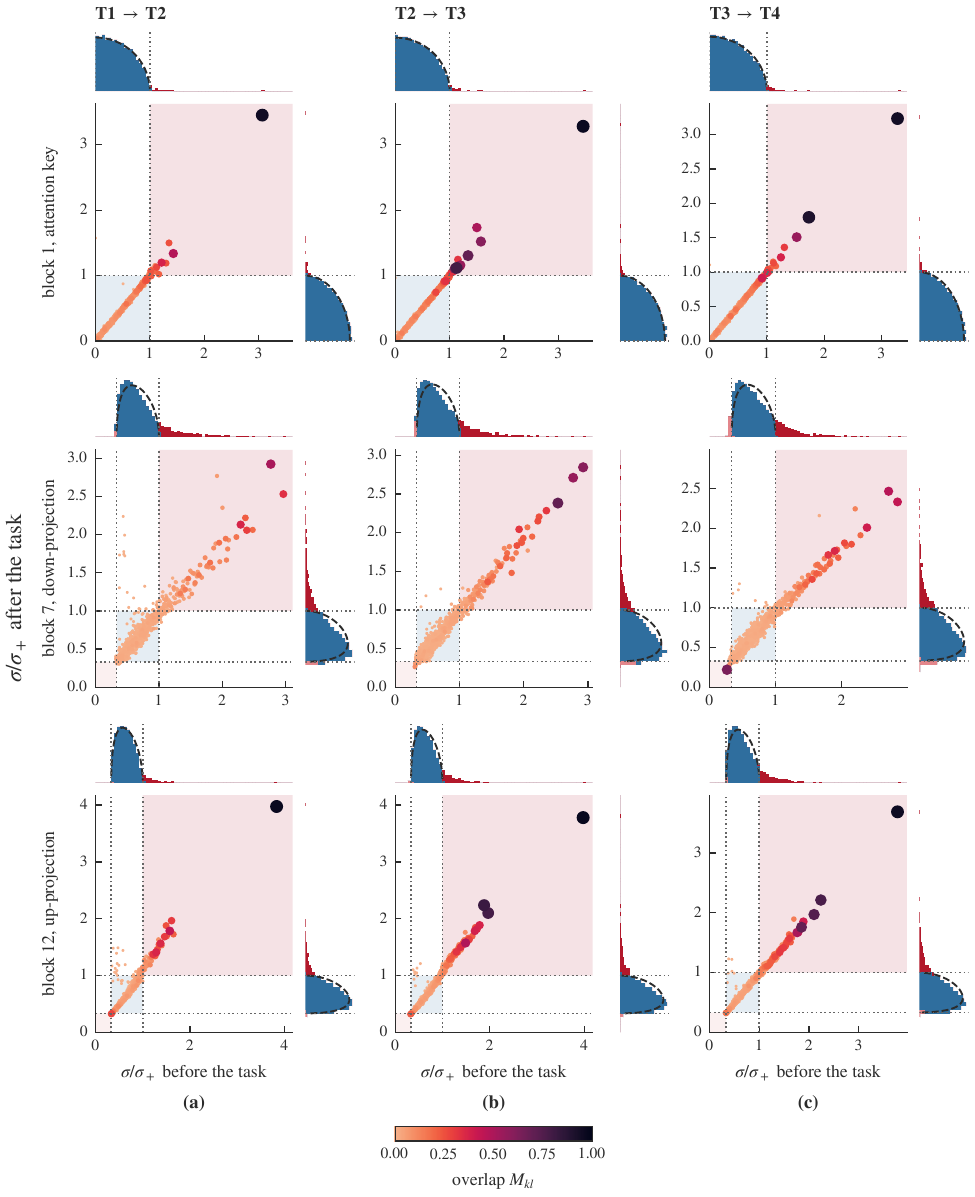}
\caption{\textbf{Matched modes in GPT-2.} As in \cref{fig:matching}, for the three inspected layers.}
\label{fig:matching-gpt2}
\end{figure}

The Hungarian assignment that tracks cohorts uses overlaps of singular vectors. \Cref{fig:matching,fig:matching-resnet,fig:matching-gpt2} place every mode at its matched partner, at their singular values relative to each checkpoint's upper MP boundary, with the spectra before and after the stage along the axes.
Across inspected layers and stages, the matched pairs cluster around the diagonal, and the median relative change in normalized singular value within a panel ranges from 1\% to 8\%.
Several modes can overlap the same mode. The persistence of \cref{sec:stratigraphy} measures each region as a subspace, without choosing a one-to-one matching.

\clearpage
\subsection{Displacement}
\label{app:displacement-results}
\label{app:displacement-depth}
\begin{figure}[H]
\centering
\includegraphics[width=\linewidth]{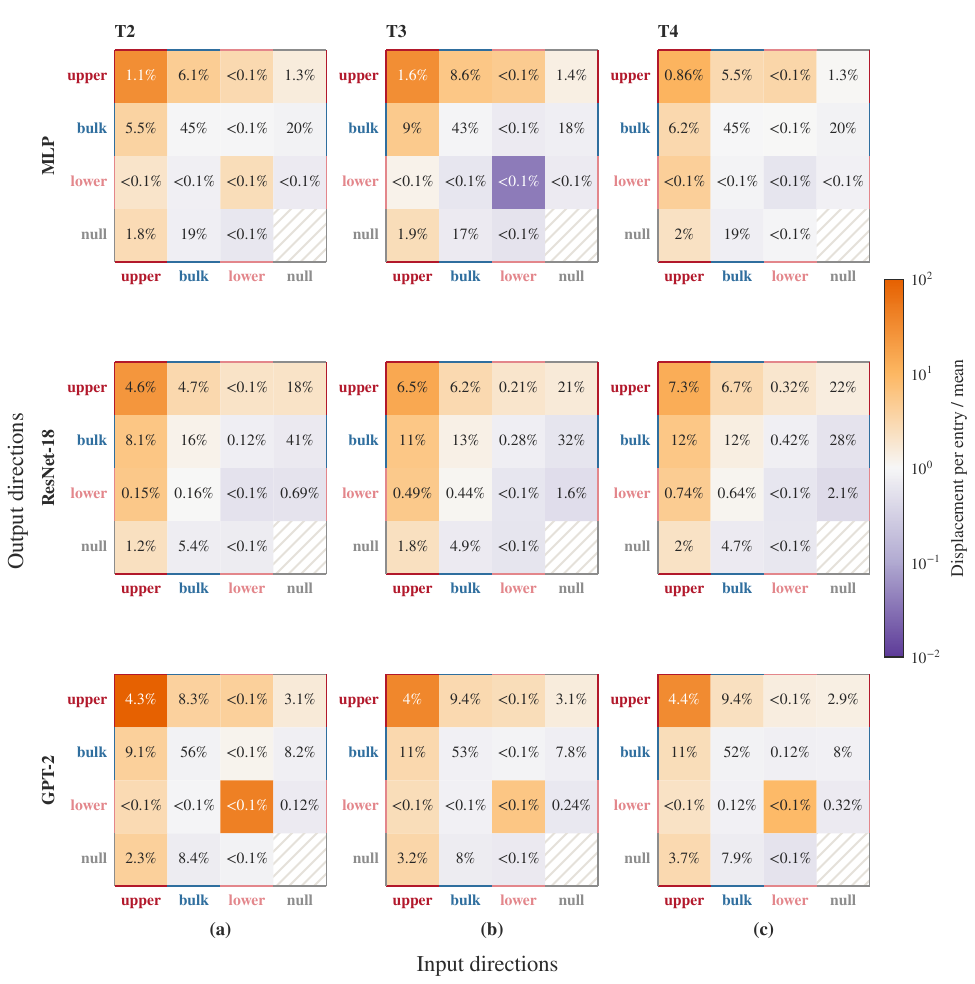}
\caption{\textbf{Displacement in the inherited singular basis.} Rows show the MLP, ResNet-18 and GPT-2, with columns for training on T2, T3 and T4. Color compares mean squared displacement per entry with the matrix-wide mean on one common logarithmic scale, with $\times1$ neutral. Numbers give each block's percentage of displacement energy, and hatching marks an empty block. Shares and enrichments are averaged over matrices, each weighing equally, and seeds (\cref{app:traffic-more}).}
\label{fig:traffic-more}
\end{figure}

\begin{figure}[H]
\centering
\includegraphics[width=\linewidth]{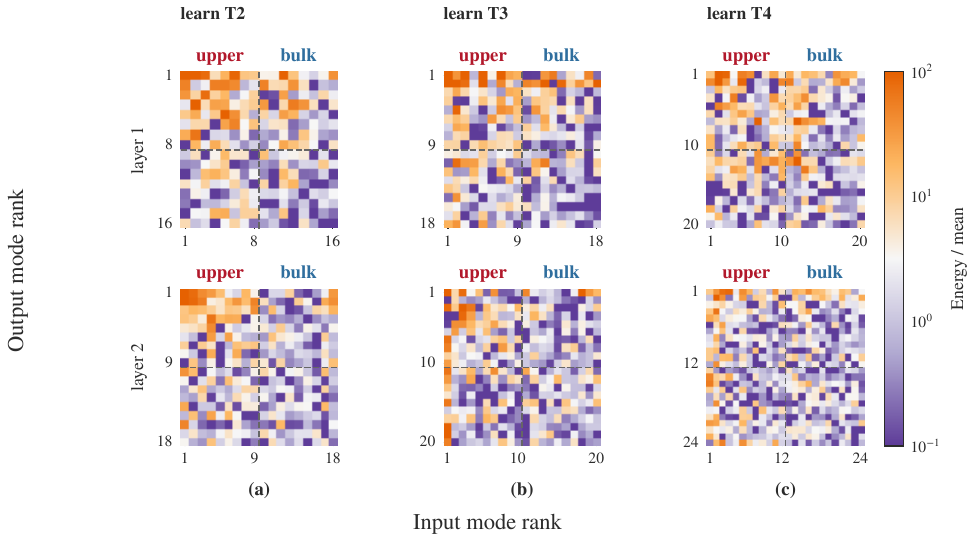}
\caption{\textbf{MLP displacement across depth and tasks.} Rows show the expanding and contracting hidden matrices; columns show training on T2, T3 and T4. Each panel shows $(u_k^\top\dW v_l)^2$ in the singular basis of the stage's starting checkpoint, relative to its mean over the matrix (\cref{app:gradient-zoom}).}
\label{fig:displacement-mlp-depth}
\end{figure}

\begin{figure}[H]
\centering
\includegraphics[width=\linewidth]{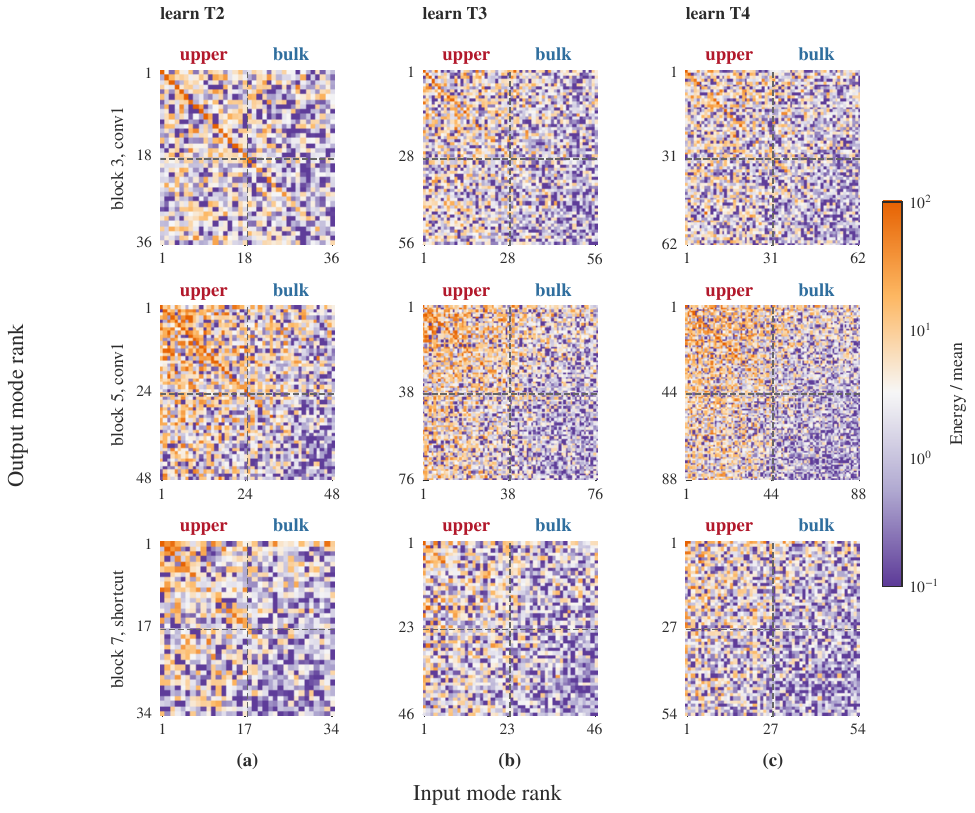}
\caption{\textbf{ResNet-18 displacement across depth and tasks.} Rows are the first convolutions of residual blocks 3 and 5 and the shortcut projection of block 7; columns show training on T2, T3 and T4. Panels are drawn as in \cref{fig:displacement-mlp-depth}.}
\label{fig:displacement-resnet-depth}
\end{figure}

\begin{figure}[H]
\centering
\includegraphics[width=\linewidth]{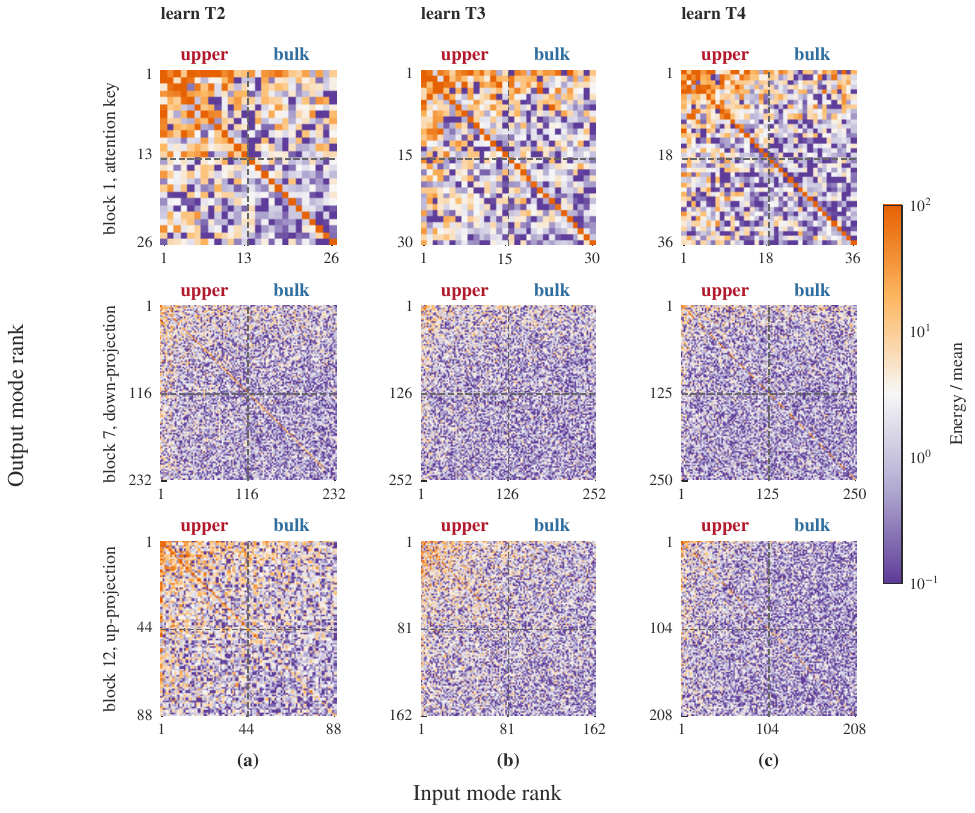}
\caption{\textbf{GPT-2 displacement across depth and tasks.} Rows are the layers of \cref{fig:esd-gpt2} and columns show training on T2, T3 and T4. Panels are drawn as in \cref{fig:displacement-mlp-depth}.}
\label{fig:displacement-gpt2-depth}
\end{figure}

We show each later task's displacement in the singular basis of its starting checkpoint, pooled over hidden matrices (\cref{fig:traffic-more}) and in every inspected layer (\cref{fig:displacement-mlp-depth,fig:displacement-resnet-depth,fig:displacement-gpt2-depth}). Across inspected layers and tasks, blocks involving upper-edge directions receive 1.5 to 5.6 times the mean displacement energy per entry of the matrix after averaging over seeds. Pooled over hidden matrices and seeds, displacement per entry in $\Delta_{EE}$ reaches 11 to 127 times the matrix average across task boundaries. In the ResNet-18 layer of \cref{fig:traffic}, the null directions span 78\% of the entries. In GPT-2, the few lower-edge modes are also enriched within their own block, which receives less than 0.1\% of the energy. For each matrix, we divide the displacement energy per entry of the block from upper-edge inputs to bulk outputs by that of the reverse block. Averaged over matrices, seeds and task boundaries, the ratio is 1.7 in the MLP, 2.0 in ResNet-18 and 1.2 in GPT-2. We compute the gradient asymmetry of \cref{sec:ports} in the same way. In the MLP, the mean combines opposite patterns in the two hidden layers. At every task boundary, after averaging over seeds, the expanding layer favors bulk inputs to upper-edge outputs and the contracting layer favors the reverse block.

\clearpage
\subsection{Incoming gradient}
\label{app:gradient-pooled}
\label{app:gradient-depth}
\begin{figure}[H]
\centering
\includegraphics[width=\linewidth]{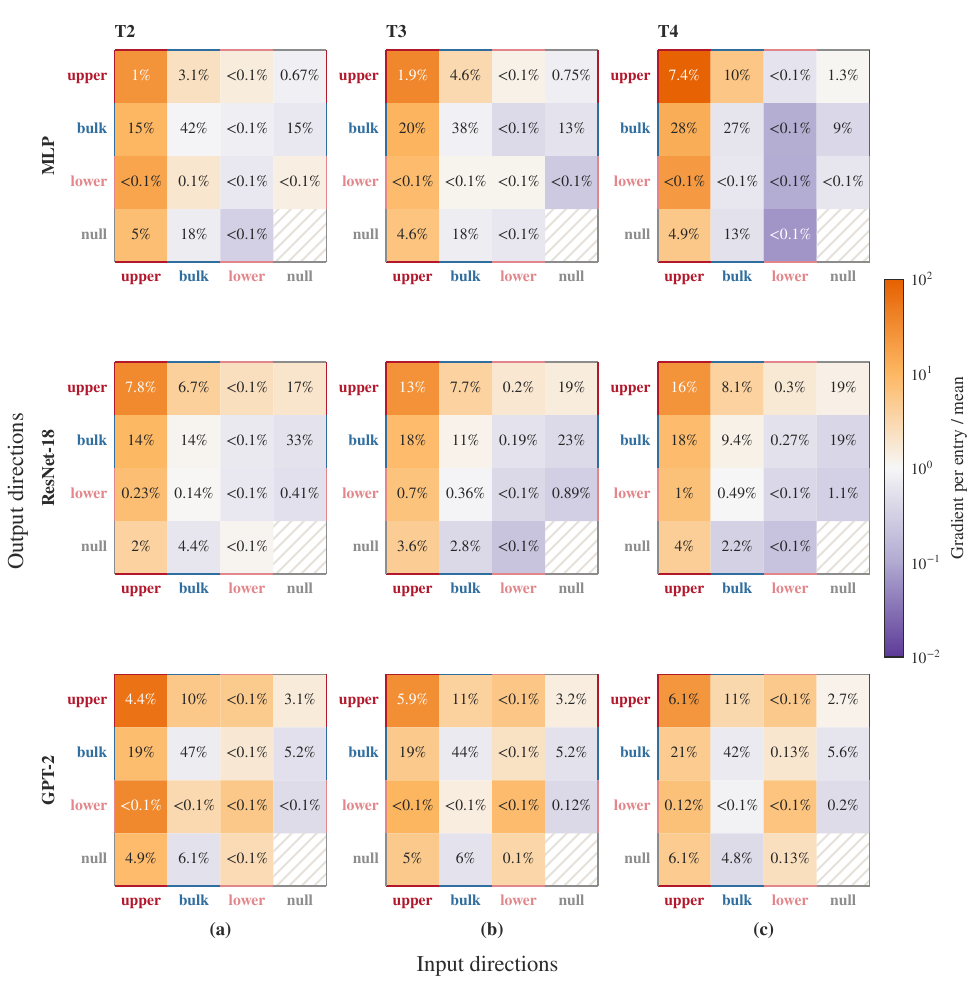}
\caption{\textbf{The incoming gradient in the inherited singular basis.} Rows show the MLP, ResNet-18 and GPT-2, and columns precede T2, T3 and T4. Blocks, colors and numbers are drawn as in \cref{fig:traffic-more}, for the gradient of the incoming task's loss at the end of the previous task.}
\label{fig:gradient-pooled}
\end{figure}

\begin{figure}[H]
\centering
\includegraphics[width=\linewidth]{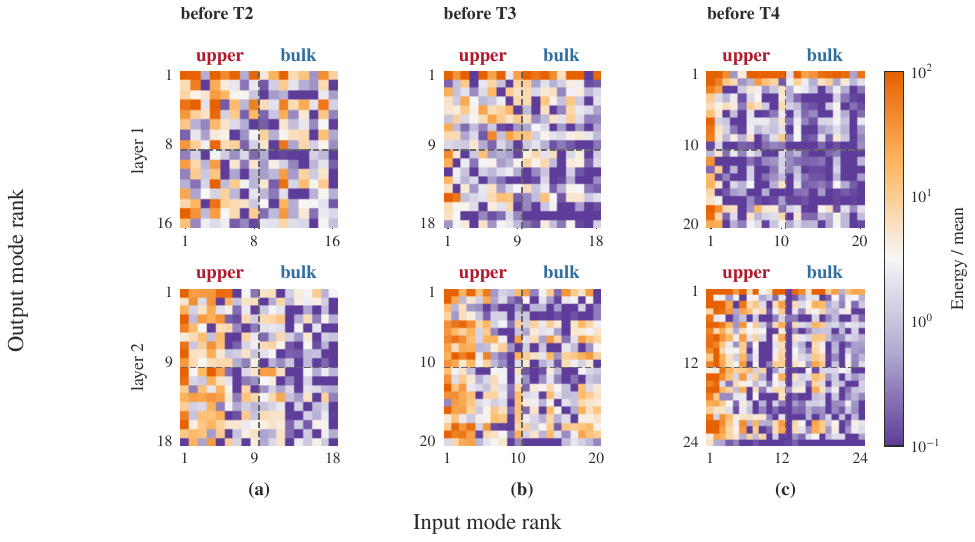}
\caption{\textbf{MLP gradients across depth and task boundaries.} Rows are the layers of \cref{fig:displacement-mlp-depth}; columns precede T2, T3 and T4. Each panel shows the squared coefficients of the incoming task's loss gradient over 1,024 examples, relative to their mean over the matrix (\cref{app:gradient-zoom}).}
\label{fig:gradient-mlp-depth}
\end{figure}

\begin{figure}[H]
\centering
\includegraphics[width=\linewidth]{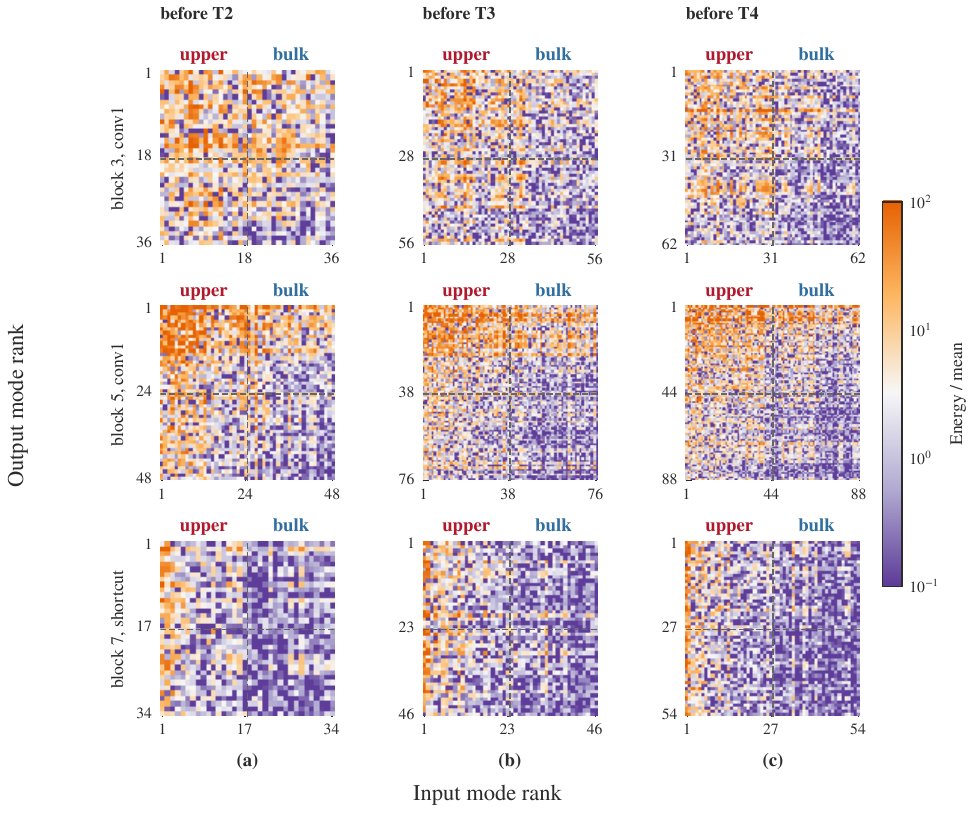}
\caption{\textbf{ResNet-18 gradients across depth and task boundaries.} Rows are the layers of \cref{fig:displacement-resnet-depth}; columns precede T2, T3 and T4. Panels are drawn as in \cref{fig:gradient-mlp-depth}.}
\label{fig:gradient-resnet-depth}
\end{figure}

\begin{figure}[H]
\centering
\includegraphics[width=\linewidth]{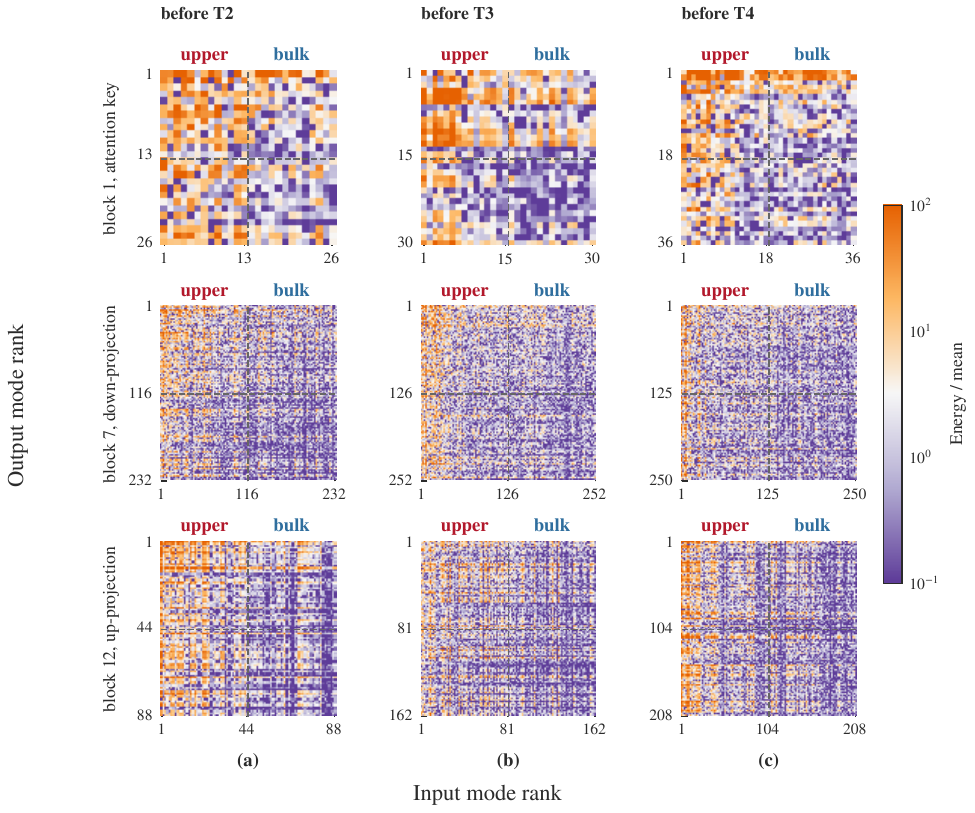}
\caption{\textbf{GPT-2 gradients across depth and task boundaries.} Rows are the layers of \cref{fig:displacement-gpt2-depth}, and columns precede T2, T3 and T4. Panels are drawn as in \cref{fig:gradient-mlp-depth}.}
\label{fig:gradient-gpt2-depth}
\end{figure}

We draw the incoming task's gradient at the end of the previous task, before any update, as the displacement of \cref{app:displacement-results}, pooled over hidden matrices (\cref{fig:gradient-pooled}) and in the same layers (\cref{fig:gradient-mlp-depth,fig:gradient-resnet-depth,fig:gradient-gpt2-depth}). In the inspected layers across all task boundaries, blocks involving upper-edge directions receive 2.2 to 8.9 times the mean gradient energy per entry of the matrix, averaged over seeds. Averaged over seeds and task boundaries, these blocks are enriched in every hidden matrix of the MLP and ResNet-18 and in 71 of 72 in GPT-2.

\clearpage
\subsection{Input and error projections across depth}
\label{app:projection-results}
At the T1-to-T2 boundary, incoming errors in the example layers of \cref{fig:ports} favor the upper edge by a factor of 2.4 to 4.3, averaged over seeds. Across the three task boundaries in GPT-2, the incoming error projections onto the upper edge are 3.0 to 7.6 times the bulk mean in the example attention key projection and 1.5 to 2.4 times in the example feed-forward up-projection. The incoming task's inputs project onto the upper edge 3.2 to 20.2 times more strongly than onto the bulk in the inspected layers, averaged over seeds. The inputs and errors of the task yet to be learned favor the upper edge in at least 90\% of hidden matrices, counted at every task boundary and in every seed (\cref{tab:incoming}). Averaged over seeds and boundaries, the incoming inputs project onto the lower edge 0.71 and 0.57 times as strongly as onto the bulk in the MLP and ResNet-18. In GPT-2, they project onto the one to four lower-edge modes of the feed-forward up-projections up to 38 times as strongly as onto the bulk, while in the down-projections the ratio stays below one.

\begin{table}[H]
\centering
\small
\caption{The incoming task's preference for the inherited upper edge before training, pooled over hidden matrices, task boundaries and seeds. Projection ratios divide the mean squared projection onto upper-edge directions by that onto bulk directions, the shares count the matrices with a ratio above one, and the asymmetry divides the gradient energy per entry of blocks mapping upper-edge inputs to bulk outputs by that of the reverse blocks.}
\label{tab:incoming}
\begin{tabular}{lccccc}
\toprule
 & \multicolumn{2}{c}{Projection ratio} & \multicolumn{2}{c}{Matrices favoring the upper edge} & Gradient \\
\cmidrule(lr){2-3}\cmidrule(lr){4-5}
 & Inputs & Errors & Inputs & Errors & asymmetry \\
\midrule
\input{tables/incoming}
\end{tabular}
\end{table}

\begin{figure}[H]
\centering
\includegraphics{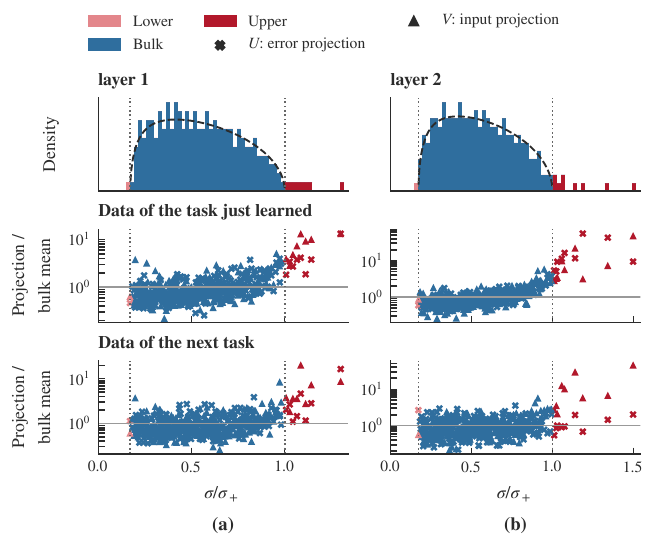}
\caption{\textbf{MLP projections across depth.} Columns are the layers of \cref{fig:displacement-mlp-depth}, drawn as in \cref{fig:ports}.}
\label{fig:ports-mlp-depth}
\end{figure}

\begin{figure}[H]
\centering
\includegraphics{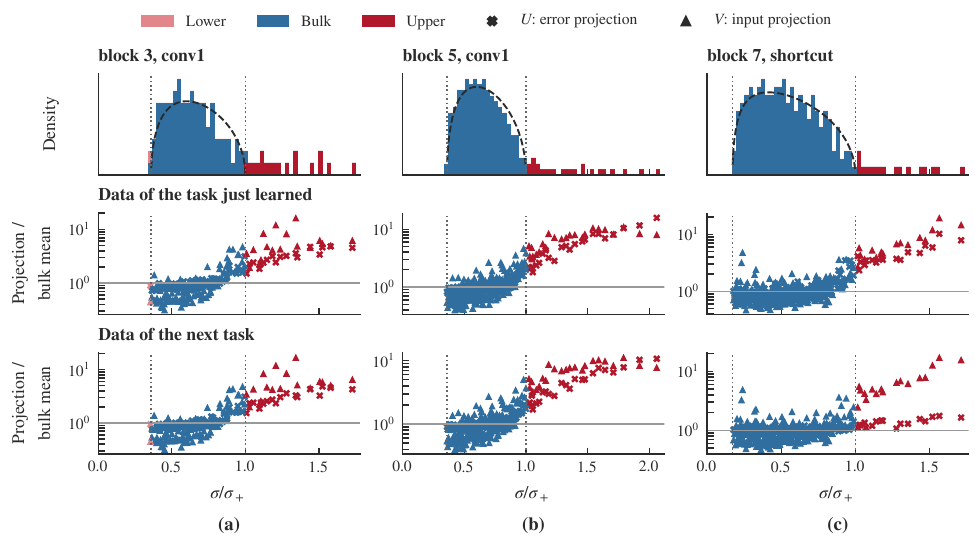}
\caption{\textbf{ResNet-18 projections across depth.} Columns are the layers of \cref{fig:displacement-resnet-depth}, drawn as in \cref{fig:ports}.}
\label{fig:ports-resnet-depth}
\end{figure}

\begin{figure}[H]
\centering
\includegraphics{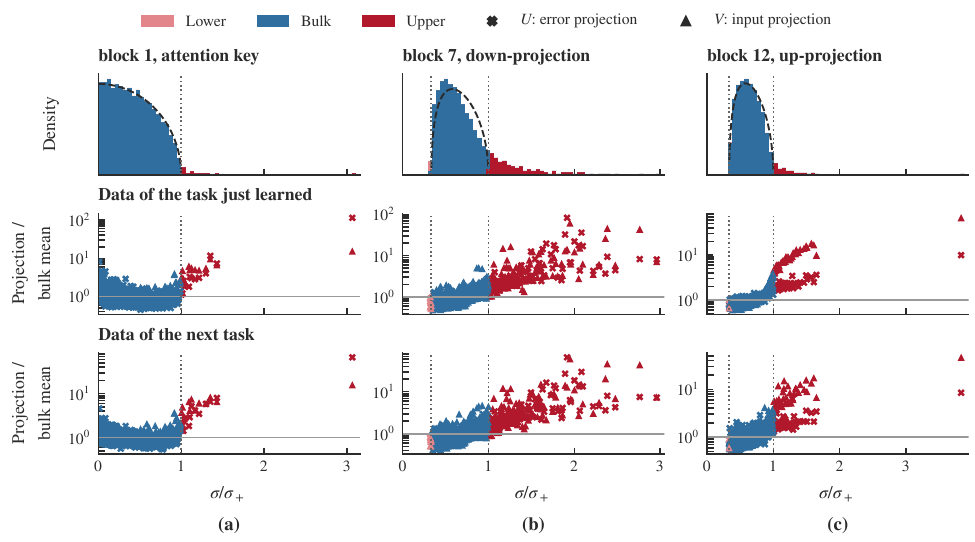}
\caption{\textbf{GPT-2 projections across depth.} Columns show the layers of \cref{fig:displacement-gpt2-depth}, drawn as in \cref{fig:ports}.}
\label{fig:ports-gpt2-depth}
\end{figure}

\clearpage
\subsection{Protection trajectories}
\label{app:retention-results}
\label{app:trajectory-results}
The trajectories follow every task through its acquisition and the later loss rise, one row per task.

\begin{figure}[H]
\centering
\includegraphics{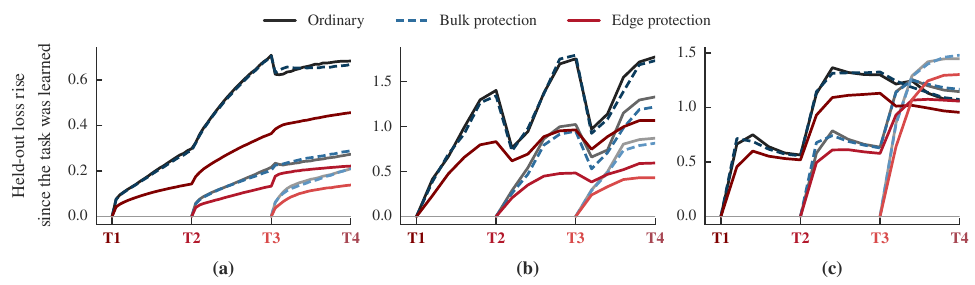}
\caption{\textbf{Protecting the edge reduces forgetting more than protecting the bulk.} Panels show \textbf{(a)}~the MLP, \textbf{(b)}~ResNet-18 and \textbf{(c)}~GPT-2. Each line follows one earlier task from the end of the stage that learned it and shows the rise of its validation loss since then, $\forget{i}{j}$, under ordinary training, bulk protection and edge protection. Bulk protection removes the bulk-involving part of as many random bulk modes (\cref{app:protection}). Lines show the mean over five seeds. \Cref{tab:protection} summarizes final forgetting and the cost to new tasks in units of learning gain.}
\label{fig:freeze-curves}
\end{figure}

\begin{figure}[H]
\centering
\includegraphics{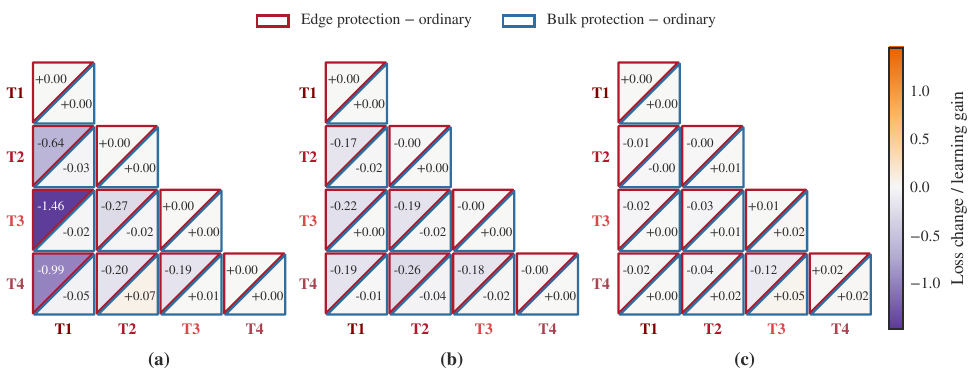}
\caption{\textbf{Edge protection against bulk protection on the training split.} Panels show \textbf{(a)}~the MLP, \textbf{(b)}~ResNet-18 and \textbf{(c)}~GPT-2, drawn as in \cref{fig:freeze} with the loss on each task's training examples.}
\label{fig:freeze-train}
\end{figure}

\begin{figure}[H]
\centering
\includegraphics{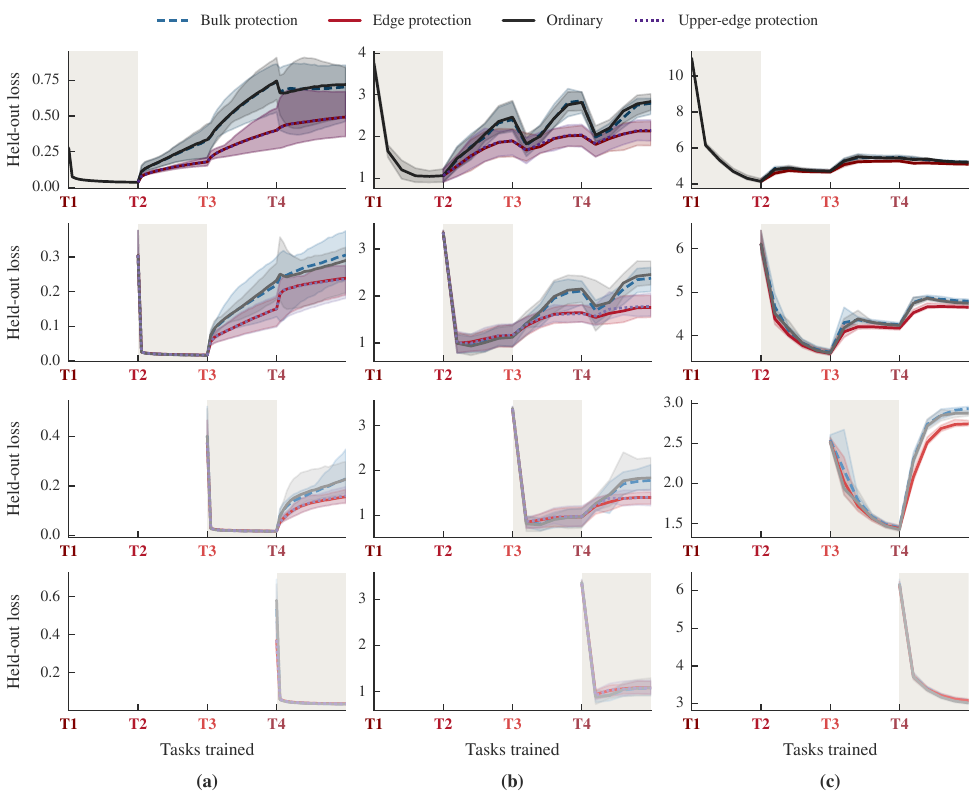}
\caption{\textbf{Full loss trajectories.} Columns show the MLP, ResNet-18 and GPT-2.
Each row follows one task, with its training interval shaded.
Lines show averages over seeds and shaded bands span the minimum and maximum.
The first task is shared, after which the training configurations proceed separately.}
\label{fig:freeze-timeline}
\end{figure}

\begin{figure}[H]
\centering
\includegraphics[width=0.49\linewidth]{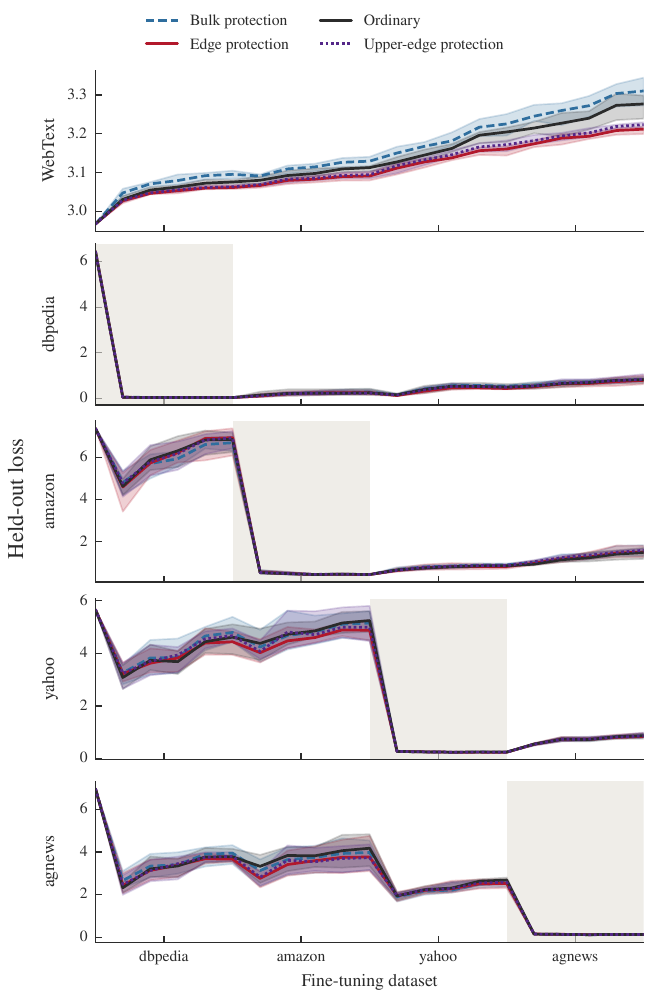}\hfill
\includegraphics[width=0.49\linewidth]{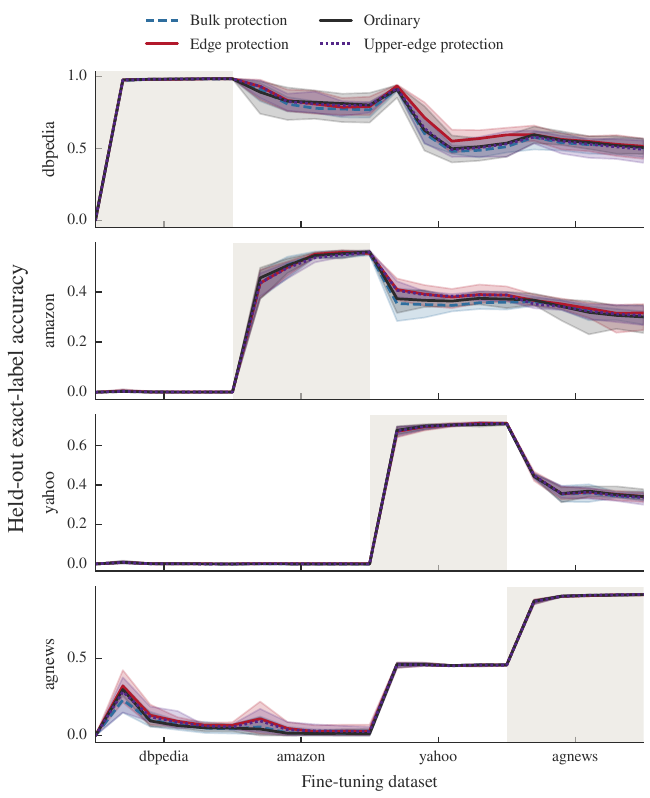}
\caption{\textbf{Full trajectories of GPT-2 fine-tuning.} Held-out loss on WebText and the four fine-tuning tasks (left) and exact-label accuracy under teacher forcing on the fine-tuning tasks (right). Each row follows one dataset under ordinary fine-tuning, bulk, upper-edge and edge protection, with its fine-tuning interval shaded where applicable. Lines show the mean over five seeds and bands their range. On each task, mean accuracy at the end of its own stage varies by less than one percentage point across configurations.}
\label{fig:gpt2-trajectories}
\end{figure}

\clearpage
\subsection{Upper-edge protection}
\label{app:upper-protection}
\begin{figure}[H]
\centering
\includegraphics[width=\linewidth]{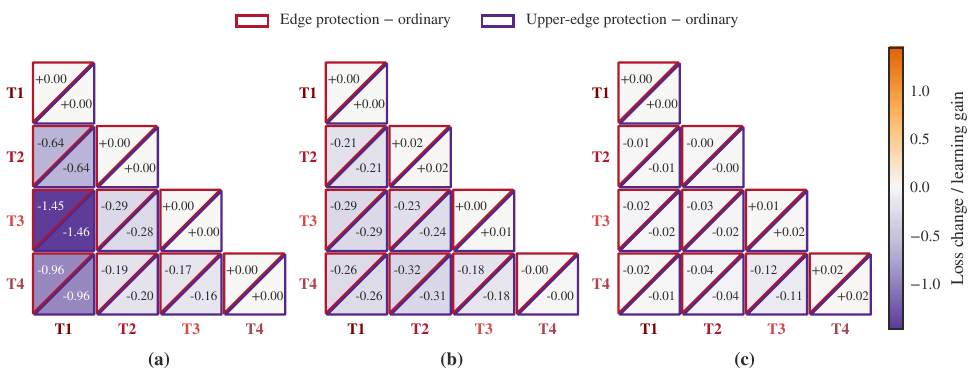}
\caption{\textbf{Upper-edge protection matches edge protection.} Panels show \textbf{(a)}~the MLP, \textbf{(b)}~ResNet-18 and \textbf{(c)}~GPT-2 trained from scratch, drawn as in \cref{fig:freeze}. In each cell, the upper-left triangle compares edge protection and the lower-right triangle upper-edge protection with ordinary training. Values average five seeds.}
\label{fig:upper-protection}
\end{figure}

\begin{figure}[H]
\centering
\includegraphics[width=\linewidth]{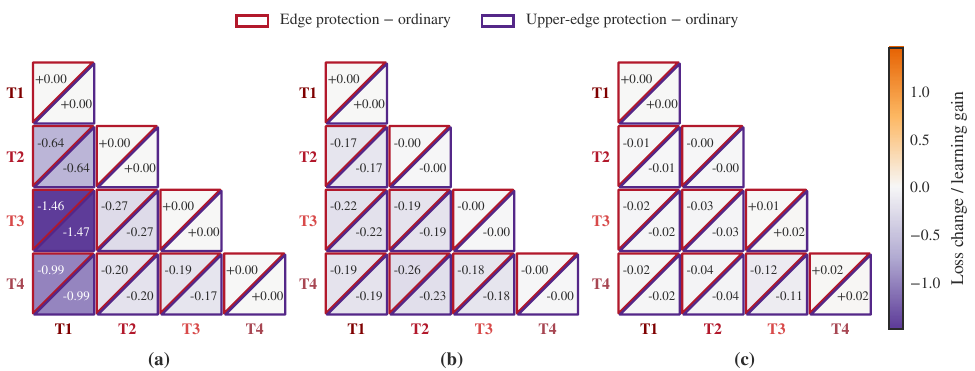}
\caption{\textbf{Upper-edge protection on the training split.} Panels show \textbf{(a)}~the MLP, \textbf{(b)}~ResNet-18 and \textbf{(c)}~GPT-2 trained from scratch, drawn as in \cref{fig:upper-protection} with the loss on each task's training examples. Values average five seeds.}
\label{fig:upper-protection-train}
\end{figure}

Protecting only the upper edge, the modes above the upper MP boundary, lowers the loss of earlier tasks about as much as protecting both edges in the MLP, ResNet-18 and GPT-2 trained from scratch, at a similar cost to new tasks (\cref{fig:upper-protection,fig:upper-protection-train}).

In GPT-2 trained from scratch, the lower edge contains only 0.8\% of hidden-matrix singular modes after the final task, compared with 9.4\% in the upper edge. Upper-edge protection alone captures most of the benefit of protecting both edges. Adding lower-edge protection yields smaller, task-dependent differences: on code, it further lowers the final training and validation losses by 1.0\% and 0.8\% of the ordinary learning gain, respectively.

In GPT-2 fine-tuning, adding the lower edge reduces the WebText loss rise from the second task onward in every seed, with a final rise of 0.24 nats per token against 0.26 under upper-edge protection. The lower and upper edges contain 2.1\% and 12.8\% of pretrained GPT-2's hidden modes, respectively.

%% file: tables/incoming.tex
MLP & 10.6 & 4.3 & 100\% & 90\% & 7.9 \\
ResNet-18 & 6.3 & 2.5 & 100\% & 100\% & 3.2 \\
GPT-2 & 12.2 & 6.8 & 100\% & 98\% & 3.4 \\
\bottomrule